\documentclass[twocolumn,letterpaper]{IEEEAerospaceCLS}  

\usepackage[]{graphicx}    
\newcommand{\ignore}[1]{}  

\usepackage{graphbox}

\usepackage{array}
\usepackage{booktabs}
\usepackage{amsmath}
\usepackage{amsfonts}
\usepackage[version=4]{mhchem}
\usepackage{siunitx}
\usepackage{longtable,tabularx}
\usepackage{mathtools}
\usepackage{multirow}

\usepackage[T1]{fontenc}\usepackage{tikz}
\usepackage{amsmath, amsfonts}
\usetikzlibrary{arrows.meta}

\usepackage[linesnumbered,ruled,vlined,norelsize,noend]{algorithm2e}
\SetInd{5pt}{10pt}
\SetNlSty{relax}{}{\enspace}

\newcommand{\bu}{\boldsymbol{u}}
\newcommand{\bx}{\boldsymbol{x}}
\newcommand{\bz}{\boldsymbol{z}}
\newcommand{\calU}{\mathcal{U}}
\newcommand{\calX}{\mathcal{X}}
\newcommand{\calM}{\mathcal{M}}

\usepackage{xcolor}
\definecolor{light-gray}{gray}{0.95}
\newcommand{\code}[1]{\colorbox{light-gray}{\texttt{#1}}}

\usepackage{hyperref}

\usepackage[font=small,skip=4pt]{caption}
\begin{document}
\title{Transformers for Trajectory Optimization \\ with Application to Spacecraft Rendezvous}

\author{%
Tommaso Guffanti$^*$\\ 
Dept. of Aeronautics \& Astronautics\\
Stanford University\\
496 Lomita Mall, Stanford, CA 94305\\
tommaso@stanford.edu
\and 
Daniele Gammelli$^*$\\
Dept. of Aeronautics \& Astronautics\\
Stanford University\\
496 Lomita Mall, Stanford, CA 94305\\
gammelli@stanford.edu
\and 
Simone D'Amico\\ 
Dept. of Aeronautics \& Astronautics\\
Stanford University\\
496 Lomita Mall, Stanford, CA 94305\\
damicos@stanford.edu
\and
Marco Pavone\\ 
Dept. of Aeronautics \& Astronautics\\
Stanford University\\
496 Lomita Mall, Stanford, CA 94305\\
pavone@stanford.edu
\thanks{\footnotesize $^*$ The first two authors have contributed equally to the paper.}
\thanks{$^\dagger$ The project’s website can be found at: \url{https://rendezvoustransformer.github.io/}}
\thanks{\footnotesize 979-8-3503-0462-6/24/$\$31.00$ \copyright2024 IEEE}              
}

\maketitle

\thispagestyle{plain}
\pagestyle{plain}

\maketitle

\thispagestyle{plain}
\pagestyle{plain}

\begin{abstract}
Reliable and efficient trajectory optimization methods are a fundamental need for autonomous dynamical systems, effectively enabling applications including rocket landing, hypersonic reentry, spacecraft rendezvous, and docking. Within such safety-critical application areas, the complexity of the emerging trajectory optimization problems has motivated the application of AI-based techniques to enhance the performance of traditional approaches. However, current AI-based methods either attempt to fully replace traditional control algorithms, thus lacking constraint satisfaction guarantees and incurring in expensive simulation, or aim to solely imitate the behavior of traditional methods via supervised learning. To address these limitations, this paper proposes the \textit{Autonomous Rendezvous Transformer} (ART)$^\dagger$ and assesses the capability of modern generative models to solve complex trajectory optimization problems, both from a forecasting and control standpoint. Specifically, this work assesses the capabilities of Transformers to (i) learn near-optimal policies from previously collected data, and (ii) warm-start a sequential optimizer for the solution of non-convex optimal control problems, thus guaranteeing hard constraint satisfaction. From a forecasting perspective, results highlight how ART outperforms other learning-based architectures at predicting known fuel-optimal trajectories. From a control perspective, empirical analyses show how policies learned through Transformers are able to generate near-optimal warm-starts, achieving trajectories that are (i) more fuel-efficient, (ii) obtained in fewer sequential optimizer iterations, and (iii) computed with an overall runtime comparable to benchmarks based on convex optimization.
\end{abstract} 
\tableofcontents

\begin{figure}[ht]
    \centering
    \includegraphics[width=\columnwidth]{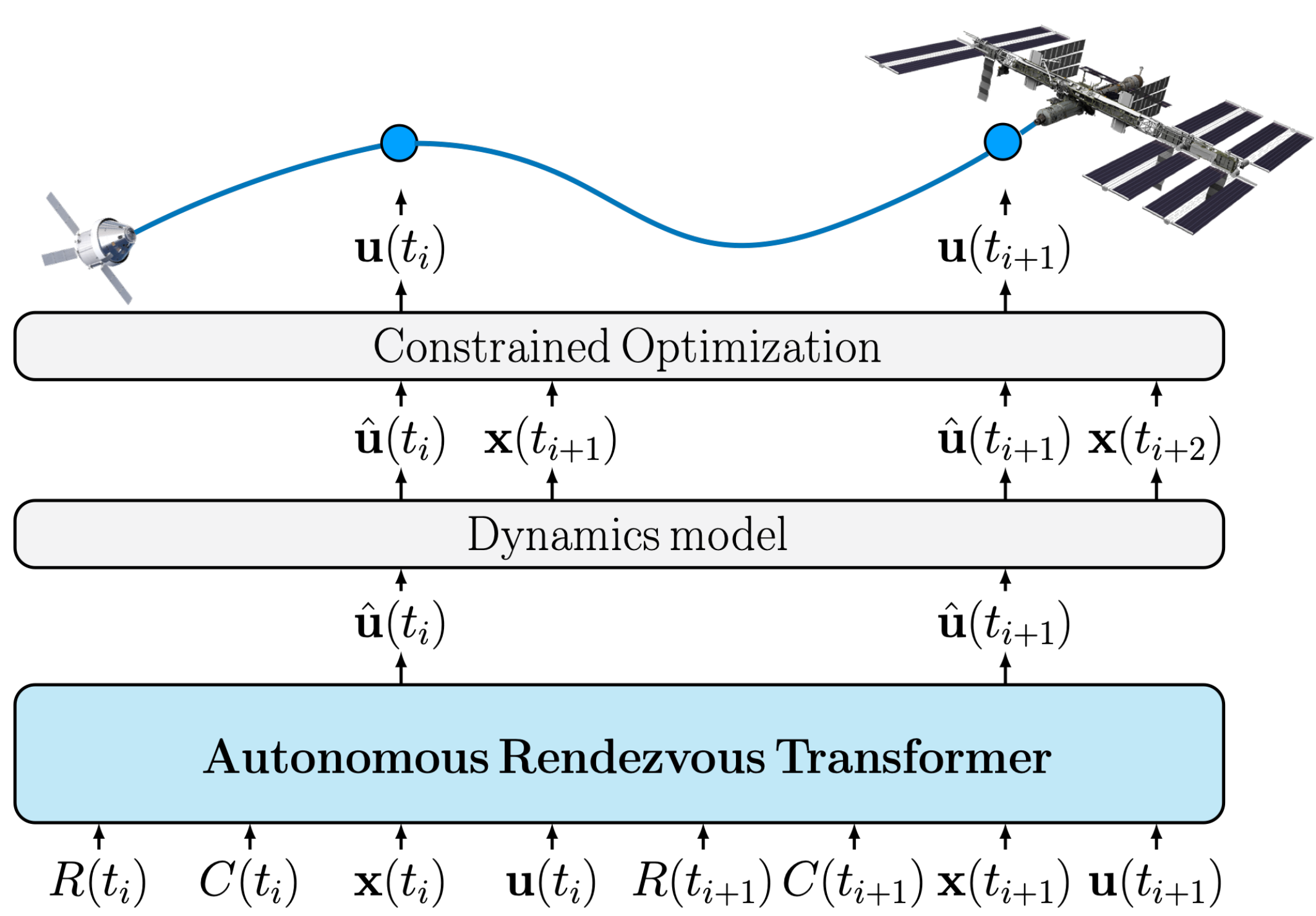}
    \caption{This paper proposes the Autonomous Rendezvous Transformer (ART) for trajectory optimization. Through the combination of efficient dynamics models and the generative capability of modern Transformer networks, the proposed framework is used to enhance traditional sequential convex programs through learning-based warm-starting, thus exploiting the advantages of data-driven approaches while ensuring hard constraint satisfaction.}
    \label{fig:fig1}
\end{figure}
\section{Introduction}
In robotics, trajectory generation is a ubiquitous approach to achieving reliable autonomy through the computation of a state and control sequence that simultaneously is dynamically feasible, satisfies constraints, and optimizes key mission objectives.
For most problems of interest, the trajectory generation problem is usually non-convex, which typically leads to formulations that are difficult to solve efficiently and reliably on-board an autonomous vehicle.
Such implications are particularly relevant for the case of spacecraft rendezvous, proximity operations, and docking (RPOD), where computational resources are forcefully limited. 
In light of this, the growing interest in exploring the application of machine learning (ML) techniques to enhance traditional trajectory generation is mainly motivated by two reasons.
First, learning-based techniques appear as promising means to approach control problems characterized by complex, multi-step, and possibly highly non-convex reward metrics \cite{silver2016mastering}.
Second, the computational overhead of using trained ML models at inference is low and possibly compatible with the limited computational capabilities available on-board spacecraft \cite{ieee_2019_computing} \cite{park2023online}.
However, the application of ML algorithms to the problem of spacecraft trajectory optimization is typically limited by the need for hard guarantees on both constraint satisfaction (e.g., dynamics feasibility, collision avoidance, path constraints) and task performance (e.g., in presence of system faults). 

Looking at the current state-of-the-art, the use of artificial intelligence (AI) for spacecraft on-board autonomy is a relatively new research area, in particular when it comes to exploring its advantages and feasibility of use for real space flight missions. 
Currently, successful deployments of autonomous capabilities within RPOD missions include unmanned technology demonstrations \cite{MEV}\cite{dennehy2011summary}\cite{damico_phd_2010} and crewed flagship-class programs \cite{d2007orion}. These missions and space vehicles did not leverage AI and still heavily relied on decisions and trajectory designs pre-approved on-the-ground, and overall had limited ability to make on-board safety-critical decisions autonomously. 
Outside of the domain of real space flight missions, the literature exploring the use of AI for spacecraft autonomy can be categorized in two groups. 
The first group tries to learn a representation for an action policy, value function, or reward model \cite{izzo2019survey}\cite{hovell_2021}\cite{FEDERICI1}\cite{FEDERICI2} by using either reinforcement learning (RL) or supervised learning (SL) techniques. 
Approaches belonging to this first group typically lack guarantees on performance and constraint satisfaction and are hindered by expensive simulation of high-fidelity spacecraft dynamics (e.g., as needed in an online RL setting). 
The second group uses learning-based components to warm-start sequential optimization solvers \cite{Banerjee_2020}\cite{cauligi2021coco}, thus being able to achieve hard constraint satisfaction while converging to a local optimum in the neighborhood of the provided warm-start.
However, while the concept of warm-starting sequential optimization solvers is appealing in principle, current approaches are typically limited in representative power (e.g., representation of a state and control trajectory via a fixed-degree polynomial), thus enforcing a limit on the set of possible applications that can be tackled.

To overcome these limitations, this paper proposes the use of Transformers to warm-start a sequential optimizer to efficiently generate near-optimal and feasible trajectories while avoiding expensive simulation (Figure \ref{fig:fig1}).
Specifically, while current approaches learn a mapping from an initial condition to a full sequence of states and controls \cite{Banerjee_2020} \cite{cauligi2021coco}, Transformers cast warm-starting as a sequence prediction problem \cite{chen2021decision} \cite{janner2021offline}, thus naturally allowing for time-dependent warm-start generation.
In Section \ref{sec_II}, this paper begins by introducing relevant literature to contextualize the proposed approach. 
In Section \ref{sec_III}, the architecture of the \textit{Autonomous Rendezvous Transformer} (ART), associated Markov Decision Process formulation, and inference pipeline are presented in the context of general constrained optimal control problems.
In Section \ref{sec_IV}, the discussion is specialized to spacecraft rendezvous, where three OCPs are introduced: one non-convex and two associated convex relaxations. 
Finally, in Section \ref{sec_V}, experimental results are presented. 
These include (i) Forecasting; i.e., the assessment of ART's capability to imitate known fuel-optimal trajectories, and (ii) Control; the assessment of ART's performance when used to warm-start sequential convex programs.

The contribution brought by this paper to the state-of-the-art is threefold\footnote{Link to the code is available on the project website$^\dagger$.}. 
\begin{itemize}
    \item A novel Transformer-based framework for the solution of general non-convex optimal control problems and its application to spacecraft rendezvous trajectory optimization.
    \item The investigation of architectural components and design decisions within the proposed framework, such as the choice of Transformer architecture and learning paradigm, MDP formulation, inference strategy, and their impact on overall performance.
    \item The demonstration through empirical analyses of ART's capability to (i) accurately forecast known fuel-optimal trajectories, and (ii) generate near-optimal warm-starts for a sequential convex program, showing advantages both in terms of achieved fuel optimality and computational efficiency.
\end{itemize}

\section{Preliminaries}\label{sec_II}
This section introduces notation and theoretical foundations underlying this work in the context of Markov decision processes and RL, and Transformers for sequence prediction.

\subsection{Markov Decision Processes and (Offline) RL}\label{sec_IIa}
Let us consider the problem of learning to control a dynamical system from experience \cite{SuttonBarto1998}.
Formally, a dynamical system is referred to as being entirely determined by a (fully-observed) Markov decision process ${\calM} = ({\calX}, {\calU}, f, \rho, {\mathcal{R}})$, where $\calX$ is a set of all possible states $\bx \in \calX$, which may be either discrete or continuous, $\calU$ is the set of possible controls $\bu \in \calU$, also discrete or continuous, $f: \calX \times \calU \rightarrow \calX $ describes the dynamics of the system, $\rho$ represents the initial state distribution $\rho(\bx(t_1))$\footnote{Throughout this work, we write $\bx(t_i)$ to denote state $\bx$ at time $t_i \in {\mathbb{R}}_{>0}$, with $i \in {\mathbb{N}}$, $i \in [1, N]$, e.g., $\bx(t_1)$ is the state at time $t_1$.}, and ${\mathcal{R}} : \calX \times \calU \rightarrow \mathbb{R}$ defines the reward function.
A trajectory consists of a sequence of states and controls $\tau = \left(\bx(t_1), \bu(t_1), \ldots,  \bx(t_N), \bu(t_N)\right)$, where $N \in \mathbb{N}$ defines the number of discrete time instants of the MDP.
The return, or reward-to-go, of a trajectory at time $t_i$, $R(t_i) = \sum_{j = i}^N {\mathcal{R}}(\bx(t_j), \bu(t_j))$, is the sum of future rewards from time $t_i$.
Intuitively, the goal of reinforcement learning is to learn a policy that maximizes the expected return ${\mathbb{E}}\left[ \sum_{i = 1}^N {\mathcal{R}}(\bx(t_i), \bu(t_i)) \right]$ in an MDP.
Most reinforcement learning algorithms follow the same basic learning loop: the agent interacts with the MDP $\mathcal{M}$ by observing some state $\bx(t_i)$, selecting a control $\bu(t_i)$, and then observing the next state $\bx(t_{i+1})$ together with a scalar reward feedback ${\mathcal{R}}(\bx(t_i), \bu(t_i))$. 
This procedure may repeat for multiple steps, during which the agent uses the observed transitions $(\bx(t_i), \bu(t_i), \bx(t_{i+1}), {\mathcal{R}}(\bx(t_i), \bu(t_i)))$ to update its policy.
Crucially, the active exploration induced by this interaction with the MDP is foundational for the success of the vast majority of reinforcement learning algorithms.
On the other hand, in offline reinforcement learning, this active interaction is precluded, and the agent is assumed to have only access to a limited offline dataset of (potentially sub-optimal) trajectories.

\begin{figure*}[ht]
    \centering
    \includegraphics[width=0.85\textwidth]{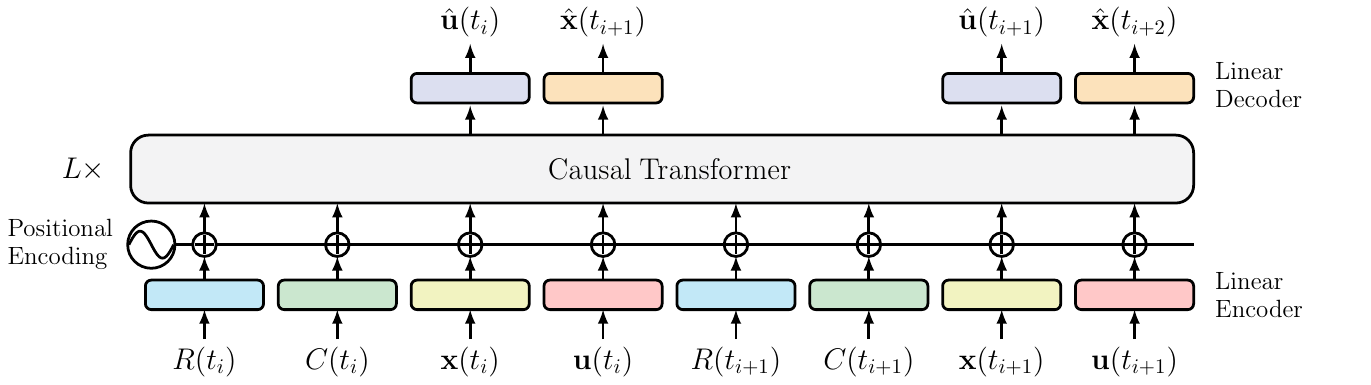}
    \caption{The Autonomous Rendezvous Transformer (ART) architecture. Each reward-to-go, constraint-to-go, state, and control vector is fed into modality-specific linear encoders, after which a positional encoding is added. The embeddings are processed by a GPT architecture which predicts future states and controls autoregressively.}
    \label{fig:ART}
\end{figure*}

\subsection{Transformers for Sequence Prediction}\label{sec_IIb}
The Transformer architecture \cite{vaswani2017attention} has recently achieved success across a wide range of applications: from natural language processing \cite{radford2018improving} and computer vision \cite{DosovitskiyEtAl2021} to robotics \cite{radosavovic2023learning}. 
At its core, the Transformer defines an architecture to efficiently model sequential data through its ability to process sequences of any length in a parallel way, as opposed to the computationally-sequential nature of e.g., recurrent neural networks.

At a high level, a Transformer \textit{network} is defined as the composition of one or more Transformer \textit{blocks}. 
Specifically, given a sequence of $N$ input embeddings $\left(\bz(t_1), \ldots, \bz(t_N)\right)$, with $\bz(t_i) \in {\mathbb{R}}^{d}$, a Transformer block is a sequence-to-sequence, dimensionality-preserving function i.e., mapping ${\mathbb{R}}^{d \times N}$ to ${\mathbb{R}}^{d \times N}$.
The core component of a Transformer block is the self-attention layer, for which the $i$-th embedding $\bz(t_i)$ is mapped via linear transformations to a key ${\boldsymbol{k}}_i \in {\boldsymbol{R}}^{d}$, query ${\boldsymbol{q}}_i \in {\mathbb{R}}^{d}$, and value ${\boldsymbol{v}}_i \in {\mathbb{R}}^{d}$ vectors.
The $i$-th output of the self-attention layer is given by weighting the values ${\boldsymbol{v}}_j$ by the normalized dot product between the query ${\boldsymbol{q}}_i$ and other keys ${\boldsymbol{k}}_j$, as
\begin{equation}\label{eq:self_attn}
    {\boldsymbol{z}}(t_i)=\sum_{j=1}^N \operatorname{softmax}\left(\left\{\left\langle {\boldsymbol{q}}_i, {\boldsymbol{k}}_{j^{\prime}}\right\rangle\right\}_{j^{\prime}=1}^N\right)_j \cdot {\boldsymbol{v}}_j,
\end{equation}
where $\langle \cdot, \cdot \rangle: {\mathbb{R}}^{d} \times {\mathbb{R}}^{d} \rightarrow {\mathbb{R}}$ represents the dot product between two vectors, $\{ \cdot \}_{j=1}^N$ is the concatenation of elements from $j=1$ through $N$, and the $\text{softmax}$ function is applied to an $N$-dimensional input vector rescaling it so that the elements of the $N$-dimensional output vector lie in the range $[0,1]$ and sum to $1$.
Intuitively, this allows the layer to assign credit, or "attend", to specific values by forming associations via similarity (i.e., large dot product) of the query and key vectors\footnote{For a broader treatment of the Transformer architecture, the reader is referred to \cite{VaswaniEtAl2017}}.
Particularly relevant for this work is the GPT architecture \cite{RadfordEtAl2019}, which modifies Eq. \eqref{eq:self_attn} with a causal self-attention mask such that elements in the sequence cannot attend to future elements in the sequence, thus respecting the temporal, or causal, structure in the sequence.
Concretely, this is achieved by replacing the summation and softmax function over $N$ with only the previous elements in the sequence ($j \in [1, i])$.


\section{Solving Constrained Optimal Control Problems using Transformers}\label{sec_III}


Let us consider the generic time-discretized non-convex constrained OCP
\begin{equation}\label{OCP}
\resizebox{0.91\linewidth}{!}{$\begin{aligned}
& \underset{\boldsymbol{u}(t_i), \boldsymbol{x}(t_i)}{\text{minimize}} & & \sum_{i = 1}^{N} J\left(\boldsymbol{x}(t_i), \boldsymbol{u}(t_i)\right) & \\
& \text{subject to} & & \boldsymbol{x}(t_{i+1}) 
= \boldsymbol{f} \left(\boldsymbol{x}(t_i), \boldsymbol{u}(t_i) \right) & \forall i \in [1, N]\\
&&& \boldsymbol{x}(t_i), \boldsymbol{u}(t_i) \in {\mathcal{C}}(t_i) & \forall i \in [1, N], \\
\end{aligned}$}
\end{equation}
where $\boldsymbol{x} \in {\mathbb{R}}^s $ is the state, $\boldsymbol{u} \in {\mathbb{R}}^a$ is the control action, $\boldsymbol{f} : {\mathbb{R}}^{s+a} \rightarrow {\mathbb{R}}^{s}$ is the non-linear dynamics, ${\mathcal{C}}$ is generic state and action constraint set, and $N \in \mathbb{N}$ defines the number of considered discrete time instants over the full OCP horizon.

\subsection{Trajectory Representation}
In order to achieve desirable trajectory generation behavior, the choice of trajectory representation should enable Transformers to (i) learn meaningful time-dependent patterns, and (ii) conditionally generate trajectories based on user-defined performance parameters.
Concretely, we define the following representation:
\begin{equation} \label{eq:traj_repr}
    \tau = \left({\mathcal{P}}(t_1), \bx(t_1), \bu(t_1), \ldots, {\mathcal{P}}(t_N), \bx(t_N), \bu(t_N)\right),
\end{equation}
where ${\mathcal{P}}$ is the set of trajectory performance parameters.
In particular, in this work, we consider two trajectory performance parameters: the reward-to-go metric $R \in {\mathbb{R}}$, directly linked to the OCP cost function as 
\begin{equation} \label{rtg}
    R(t_i) = \sum_{j = i}^{N}  {\mathcal{R}}(\bx(t_j), \bu(t_j)) =  - \sum_{j = i}^{N}{J(\boldsymbol{x}(t_j), \boldsymbol{u}(t_j))},
\end{equation}
and the constraint-to-go metric $C \in {\mathbb{N}}$, directly linked to the OCP constraint set as 
\begin{equation}\label{ctg}
    C(t_i) = \sum_{j = i}^{N}{{\textsf{C}}(t_j)}, 
\end{equation}
where ${\textsf{C}}(t_j)$ checks for constraint violation as
\begin{equation}\label{ctg2}
    {\textsf{C}}(t_j) = 
\begin{cases}
    1, & \text{if } \exists  \boldsymbol{x}(t_j), \boldsymbol{u}(t_j) \notin {\mathcal{C}}(t_j) \\
    0, & \text{otherwise},
\end{cases}
\end{equation}
thus allowing the Transformer to generate trajectories based on \textit{future} desired returns and constraint violations.
Specifically, through such a representation, the user is able to specify a desired performance at inference (i.e., reward-to-go $R(t_1)$ and constraint-to-go $C(t_1)$), as well as the initial state $\bx(t_1)$, to initiate the sequential generation of the trained Transformer.


\subsection{Autonomous Rendezvous Transformer Architecture}
Given an input trajectory represented as in Eq.\eqref{eq:traj_repr} and a pre-specified maximum context length $K$, ART receives the last $K$ timesteps, for a total of $4K$ sequence elements, one for each modality: reward-to-go, constraint-to-go, state, and control.
To obtain a sequence of $4K$ embeddings, each element is projected through a modality-specific linear transformation, or encoder, to the embedding dimension.
Moreover, as in \cite{chen2021decision}, an embedding for each timestep is learned and added to each element embedding. 
The resulting sequence of embeddings is then processed by a GPT model, which predicts future states and controls through autoregressive generation.
Figure \ref{fig:ART} provides a schematic illustration.

\subsection{Training}

Given an offline dataset of trajectories obtained through the solution of Problem \eqref{OCP}, the first step is to reorganize the trajectories to be consistent with the representation in Eq. \eqref{eq:traj_repr} in order to obtain a dataset amenable for Transformer training of the form:
\begin{equation}
\resizebox{0.91\linewidth}{!}{$
\begin{gathered}
\left(\begin{array}{ccccccc}
{\mathcal{P}}^1(t_1) & \boldsymbol{x}^1(t_1) & \boldsymbol{u}^1(t_1) & \ldots & {\mathcal{P}}^1(t_N) & \boldsymbol{x}^1(t_N) & \boldsymbol{u}^1(t_N) \\
\vdots & \vdots & \vdots & & \vdots & \vdots & \vdots \\
{\mathcal{P}}^{N_{d}}(t_1) & \boldsymbol{x}^{N_{d}}(t_1) & \boldsymbol{u}^{N_{d}}(t_1) & \ldots & {\mathcal{P}}^{N_{d}}(t_N) & \boldsymbol{x}^{N_{d}}(t_N) & \boldsymbol{u}^{N_{d}}(t_N) \\
\end{array}\right),
\end{gathered}$}
\end{equation}
where each row represents a trajectory, $N_d \in {\mathbb{N}}$ is the number of data trajectories, and ${\mathcal{P}} = \left\{R, C\right\}$.
Training is performed with the teacher-forcing procedure \cite{GoyalEtAl2016}, as usually done to train sequence models.
Specifically, denoting the parameters of ART as $\theta$ and $||\cdot||_2$ as the L2-norm, ART is trained via the minimization of the following squared-error loss function:
\begin{equation}
\resizebox{0.95\linewidth}{!}{$\begin{aligned}\label{eq:art_loss}
    {\mathcal{L}}(\tau) = \sum_{n=1}^{N_d}\sum_{i=1}^{N} & \left(||\bx^n(t_i) - \hat \bx^n(t_i) ||_2^2 + ||\bu^n(t_i) - \hat \bu^n(t_i) ||_2^2 \right)\\
    & \hat \bx^n(t_i) = \text{ART}_{\theta}\left(\tau^n_{<t_i}\right)\\
    & \hat \bu^n(t_i) = \text{ART}_{\theta}\left(\tau^n_{<t_i}, \bx^n(t_i)\right),
\end{aligned}$}
\end{equation}
where $\text{ART}_{\theta}(\cdot)$ denotes one-step prediction via ART, $\tau_{<t_i}$ represents a trajectory from timesteps 1 through $i-1$, and $\hat \bx(t_i)$, $\hat \bu(t_i)$ define a predicted state and control, respectively.

\subsection{Trajectory Generation Approaches}

Once trained, ART can be used to generate any element of the trajectory given the context composed by the previous elements in the sequence. 
Therefore, ART can be used to generate both future actions (i.e., as a controller), future states (i.e., as an approximate dynamics model), and even, in principle, the associated performance parameters (e.g., approximate future reward, i.e., value function). 
This work investigates the capability of ART to infer good actions, which, when applied in the dynamics of the system, produce a dynamically feasible, high-performance trajectory. 
Specifically, given an inferred control $\hat \bu(t)$, this work investigates two approaches to compute future states: (i) \textit{trasformer-only}, and (ii) \textit{dynamics-in-the-loop}. 
The former directly infers states through the Transformer, while the latter directly uses the known dynamics $\boldsymbol{f}\left(\bx(t), \hat \bu(t)\right)$. 
If on one hand generating the next state through $\boldsymbol{f}\left(\cdot\right)$ enforces dynamics feasibility by construction, using the Transformer gives no guarantee of constraint satisfaction.
In both cases, after executing the generated control for the current state, the proposed inference strategy (i) decreases the reward-to-go $R(t)$ by the instantaneous reward (negative cost), (ii) decreases the constraint-to-go $C(t)$ by the instantaneous constraint violation ${\textsf{C}}(t)$, and (iii) repeats until the horizon is reached.

The investigated inference pipelines are described in Algorithms \ref{Algo_inf_tonly} and \ref{Algo_inf_dyn}, and represented graphically in Figure \ref{fig:fig1}.

\IncMargin{1em}
\begin{algorithm}[t]\small
\caption{ART inference - Transformer-only}\label{Algo_inf_tonly}
    \DontPrintSemicolon
    \SetAlgoLined
    \KwIn{$R(t_1), C(t_1), \boldsymbol{x}(t_1)$} 
    \KwOut{$\boldsymbol{x}(t_i), \hat{\boldsymbol{u}}(t_i)$, $\forall i \in [1, N]$}
    \KwData{${\mathrm{ART}}(.)$}
    \Begin{ 
    $\hat{\boldsymbol{x}}(t_1) = \boldsymbol{x}(t_1)$ \;
    ${\mathrm{ctx}}_{u}(t_{1}) = \{ R(t_1), C(t_1), \hat{\boldsymbol{x}}(t_1) \}$ \;
    \For{$i = 1$ \KwTo $N$}{
    $\hat{\boldsymbol{u}}(t_{i}) = {\mathrm{ART}}\left( {\mathrm{ctx}}_{u}(t_{i}) \right)$ \;
    \If{$i < N$}{
    $R(t_{i+1}) = R(t_{i}) - \left(- J\left(\hat{\boldsymbol{x}}(t_{i}), \hat{\boldsymbol{u}}(t_{i}) \right) \right)$\;
    \eIf{$\hat{\boldsymbol{x}}(t_i), \hat{\boldsymbol{u}}(t_i) \notin {\mathcal{C}}(t_i)$}{
    ${\textsf{C}}(t_i) = 1$\;
    }{
    ${\textsf{C}}(t_i) = 0$\;
    }
    $C(t_{i+1}) = C(t_{i}) - \textsf{C}(t_{i})$ \;
    ${\mathrm{ctx}}_{x}(t_{i+1}) = \{ R(t_j), C(t_j), \hat{\boldsymbol{x}}(t_k), \hat{\boldsymbol{u}}(t_k) \}$, $\forall j \in [1, i+1]$, $\forall k \in [1, i]$ \;
    $\hat{\boldsymbol{x}}(t_{i+1}) = {\mathrm{ART}}\left( {\mathrm{ctx}}_{x}(t_{i+1}) \right)$ \;
    ${\mathrm{ctx}}_{u}(t_{i+1}) = \{ R(t_j), C(t_j), \hat{\boldsymbol{x}}(t_j), \hat{\boldsymbol{u}}(t_k) \}$, $\forall j \in [1, i+1]$, $\forall k \in [1, i]$ \;
    $\boldsymbol{x}(t_{i+1}) = \boldsymbol{f} \left(\boldsymbol{x}(t_i), \hat{\boldsymbol{u}}(t_i) \right)$ \;   
    }   
    }
    \Return $\boldsymbol{x}(t_i), \hat{\boldsymbol{u}}(t_i)$, $\forall i \in [1, N]$ \;
    }
\end{algorithm}
\DecMargin{1em}

\IncMargin{1em}
\begin{algorithm}[t]\small
\caption{ART inference - Dynamics-in-the-loop}\label{Algo_inf_dyn}
    \DontPrintSemicolon
    \SetAlgoLined 
    \KwIn{$R(t_1), C(t_1), \boldsymbol{x}(t_1)$} 
    \KwOut{$\boldsymbol{x}(t_i), \hat{\boldsymbol{u}}(t_i)$, $\forall i \in [1, N]$}
    \KwData{${\mathrm{ART}}(.)$}
    \Begin{ 
    ${\mathrm{ctx}}_{u}(t_{1}) = \{ R(t_1), C(t_1), \boldsymbol{x}(t_1) \}$ \;
    \For{$i = 1$ \KwTo $N$}{
    $\hat{\boldsymbol{u}}(t_{i}) = {\mathrm{ART}}\left( {\mathrm{ctx}}_{u}(t_{i}) \right)$ \;
    \If{$i < N$}{
    $R(t_{i+1}) = R(t_{i}) - \left(- J\left(\boldsymbol{x}(t_{i}), \hat{\boldsymbol{u}}(t_{i}) \right) \right)$\;
    \eIf{$\boldsymbol{x}(t_i), \hat{\boldsymbol{u}}(t_i) \notin {\mathcal{C}}(t_i)$}{
    ${\textsf{C}}(t_i) = 1$\;
    }{
    ${\textsf{C}}(t_i) = 0$\;
    }
    $C(t_{i+1}) = C(t_{i}) - \textsf{C}(t_{i})$ \;
    $\boldsymbol{x}(t_{i+1}) = \boldsymbol{f} \left(\boldsymbol{x}(t_i), \hat{\boldsymbol{u}}(t_i) \right)$ \;
    ${\mathrm{ctx}}_{u}(t_{i+1}) = \{ R(t_j), C(t_j), \boldsymbol{x}(t_j), \hat{\boldsymbol{u}}(t_k) \}$, $\forall j \in [1, i+1]$, $\forall k \in [1, i]$ \;
    }
    }   
    \Return  $\boldsymbol{x}(t_i), \hat{\boldsymbol{u}}(t_i)$, $\forall i \in [1, N]$ \;
    }
\end{algorithm}
\DecMargin{1em}

\subsection{Practical Considerations}
The input parameters of Algorithms \ref{Algo_inf_tonly} and \ref{Algo_inf_dyn} are fundamentally three: (i) the initial state $\boldsymbol{x}(t_1)$ which is fixed by the problem scenario, (ii) the initial reward-to-go $R(t_1)$, which quantifies the expected optimality level (negative total cost) to be achieved by the generated trajectory, and (iii) the initial constraint-to-go $C(t_1)$, which quantifies the expected feasibility level (costraint satisfaction) to be achieved by the generated trajectory. $R(t_1)$ and $C(t_1)$ are design parameters that can be used to condition the trajectory generation. 
In this work, we propose to select $R(t_1)$ as a (negative) quantifiable lower bound of the optimal cost and $C(t_1) = 0$ to condition the generation of near-optimal and feasible trajectories.

\section{Application to Spacecraft Rendezvous}\label{sec_IV}

In the considered scenario, a servicer spacecraft has to rendezvous and dock with a target spacecraft or target space station. The target lies on an absolute reference orbit uniquely defined by a set of orbital elements (OE): $\textbf{\oe} \in {\mathbb{R}}^6$. Let us use the quasi non-singular OE definition \cite{Vallado}: $\textbf{\oe} = \left\{ a, \nu, e_{x}, e_{y}, i, \Omega \right\}$, with $a$ the semi-major axis, $\nu = M + \omega$ the mean argument of latitude, $M$ the mean anomaly, $\omega$ the argument of periapsis, $\{e_x, e_y \} = \{ e \cos(\omega), e \sin(\omega) \}$ the eccentricity vector, $e$ the eccentricity, $i$ the inclination, and $\Omega$ the right ascension of the ascending node. The relative orbital motion of the servicer with respect to the target can be expressed equivalently using a relative Cartesian state, or using Relative Orbital Elements (ROE) which are nonlinear combinations of the OE of the servicer and the target and equivalent to the integration constants of the linearized equations of relative orbital motion \cite{damico_phd_2010}\cite{damico_ROEIC_2005}. The relative Cartesian state is expressed in the Radial Tangential Normal (RTN) reference frame centered on the target, and it is defined as: $\delta \boldsymbol{\chi} = \{\delta \boldsymbol{r}, \delta \boldsymbol{v} \} \in {\mathbb{R}}^6$ where $\delta \boldsymbol{r} = \{\delta r_r, \delta r_t, \delta r_n \} \in {\mathbb{R}}^3$ is the relative position and $\delta \boldsymbol{v} = \{\delta v_r, \delta v_t, \delta v_n \} \in {\mathbb{R}}^3$ is the relative velocity. The quasi-nonsingular ROE state \cite{damico_phd_2010} is a nonlinear combination of the OE of the servicer (noted with subscript $s$) and the target as: $\delta \textbf{\oe} = \{ \delta a, \delta \lambda, \delta e_x, \delta e_y, \delta i_x, \delta i_y \} \in {\mathbb{R}}^6$, where $\delta a = (a_s - a)/a$ is the relative semi-major axis, $\delta \lambda = \nu_s - \nu + \left(\Omega_s - \Omega \right) \cos(i)$ the relative mean longitude, $\{\delta e_x, \delta e_y \} = \delta e \{\cos(\varphi), \sin(\varphi) \} = \{e_s \cos(\omega_s) - e \cos(\omega), e_s \sin(\omega_s) - e \cos(\omega)\}$ the relative eccentricity vector, and $\{\delta i_x, \delta i_y \}  = \delta i \{\cos(\phi), \sin(\phi) \} = \{i_s - i, \left(\Omega_s - \Omega \right) \sin(i)\}$ the relative inclination vector. The relative Cartesian and ROE state have a first-order one-to-one mapping
\begin{equation}
\delta \boldsymbol{\chi}(t) \approx \boldsymbol{\Psi}(t) \delta \textbf{\oe}(t),
\end{equation}
with $\boldsymbol{\Psi} \in {\mathbb{R}}^{6\times 6}$ defined in \cite{damico_phd_2010}\cite{sullivan_nonlinear_2017}\cite{guffanti_jgcd_2023} for both near-circular and eccentric orbits.

The nonintegrable perturbed and controlled time-discretized relative orbital dynamics of the servicer with respect to the target can be expressed favorably on the ROE state using variation of parameters \cite{Vallado} as
\begin{equation}\label{dynamics}
 \delta \textbf{\oe}(t_{i+1}) = \boldsymbol{\Phi}(\delta t, t_i) \delta \textbf{\oe}(t_i) + \boldsymbol{\Phi}(\delta t, t_i) \boldsymbol{B}(t_i) \Delta \boldsymbol{v}(t_i),
\end{equation}
where $\boldsymbol{\Phi} \in {\mathbb{R}}^{6\times 6}$ is a state transition matrix including a variety of orbital perturbations relevant in different orbit scenarios \cite{koenig_2017_new}\cite{guffanti_linear_2019}\cite{guffanti_long-term_2017}, $\boldsymbol{B} \in {\mathbb{R}}^{6\times 3}$ is the control input matrix defined in \cite{gaias_imp_2015}\cite{chernick_2018_closed} for both near-circular and eccentric orbits, and $\Delta \boldsymbol{v} \in {\mathbb{R}}^3$ is delta-velocity applied by the servicer. 

In the rendezvous scenario, some constraints are more naturally formulated on the relative Cartesian state (e.g., terminal docking conditions and waypoints, approach cones, etc.) \cite{pinglu_rpod}\cite{malyuta_rpod}\cite{malyuta_scp_2022}, whereas others are more advantageously formulated on the ROE state (e.g., perturbed orbital dynamics, passive safety conditions and waypoints, etc.) \cite{koenig_2017_new}\cite{damico_proximity_2006}\cite{guffanti_jgcd_2023}. At small spacecraft separation, the defined first-order linear map $\boldsymbol{\Psi}$ allows to accurately map constraints from one state representation to the other while retaining their possible convexity. In the following, given the dynamics models used, we employ a ROE representation in the formalization of the optimal control problems. 

\subsection{Optimal Control Problems}
In this paper, we apply the proposed modeling framework to three optimal control problems (OCPs) of relevance for the considered rendezvous scenario. These are: (1) a convex two-point-boundary-value-problem (TPBVP) between an initial and final condition, (2) a convex rendezvous problem including a pre-docking waypoint and an approach cone constraint, and (3) a non-convex rendezvous problem including also a keep-out-zone constraint around the target. While the first two problems can be solved directly using convex optimization tools \cite{ecos_13}, the latter requires the formalization of a Sequential Convex Program (SCP) \cite{scp_2000} \cite{liu_scp_2014} \cite{malyuta_scp_2022}. In the remainder of the paper, the capability of ART to generate effective warm-starting trajectories for this SCP is assessed and benchmarked against the performances obtained by using the solutions of the convex problems as warm-starts.

\subsubsection{Problem 1: Two Point Boundary Value Problem} 
The time-discretized fuel-optimal two-point-boundary-value-problem is formalized as
\begin{equation}
\label{OCP_TPBVP}
\resizebox{0.98\linewidth}{!}{$\begin{aligned}
& \underset{\Delta \boldsymbol{v}(t_i), \delta \textbf{\oe}(t_i)}{\text{minimize}} & & \sum_{i = 1}^{N} || \Delta \boldsymbol{v}(t_i) ||_2 &\\
& \text{subject to} & & \delta \textbf{\oe}(t_{i+1}) = \boldsymbol{\Phi}(\delta t, t_i) \delta \textbf{\oe}(t_i) + \boldsymbol{\Phi}(\delta t, t_i) \boldsymbol{B}(t_i) \Delta \boldsymbol{v}(t_i) & \forall i \in [1, N]\\
&&& \delta \textbf{\oe}(t_1) \equiv \textrm{state estimate} & \\
&&& \delta \textbf{\oe}(t_{N+1}) \equiv \textrm{terminal guidance}, & \\
\end{aligned}$}
\end{equation}
The initial condition is the most updated state estimate provided by the navigation filter and can represent (as in the next section) an initial passively safe trajectory around the target. The terminal condition is provided by the on-board guidance and can represent either a relative orbit correction, a reconfiguration way-point, or (as in the next section) a docking port. To minimize fuel consumption, the minimization of the sum of the L2-norms of the applied $\Delta \boldsymbol{v}$ is sought.

\subsubsection{Problem 2: Convex Rendezvous Problem}

The time-discretized fuel-optimal convex rendezvous problem is formalized as
\begin{equation}
\label{OCP_RPOD_CVX}
\resizebox{0.98\linewidth}{!}{$\begin{aligned}
& \underset{\Delta \boldsymbol{v}(t_i), \delta \textbf{\oe}(t_i)}{\text{minimize}} & & \sum_{i = 1}^{N} || \Delta \boldsymbol{v}(t_i) ||_2 &\\
& \text{subject to} & & \delta \textbf{\oe}(t_{i+1}) = \boldsymbol{\Phi}(\delta t, t_i) \delta \textbf{\oe}(t_i) + \boldsymbol{\Phi}(\delta t, t_i) \boldsymbol{B}(t_i) \Delta \boldsymbol{v}(t_i) & \forall i \in [1, N]\\
&&& \delta \textbf{\oe}(t_1) \equiv \textrm{initial passively safe relative orbit} & \\
&&& \boldsymbol{\Psi}(t_{N_{w.p.}}) \delta \textbf{\oe}(t_{N_{w.p.}}) = \{ \delta \boldsymbol{r}_{w.p.}, \boldsymbol{0}_{3\times 1} \} & \\
&&& || \boldsymbol{A}_{a.c.}(t_i) \delta \textbf{\oe}(t_{i}) + \boldsymbol{b}_{a.c.} ||_2 \leq \boldsymbol{c}^T_{a.c.}(t_i) \delta \textbf{\oe}(t_{i}) + d_{a.c.} & \forall i \in  [N_{w.p.}, N] \\
&&& \boldsymbol{\Psi}(t_{N+1}) \delta \textbf{\oe}(t_{N+1}) = \{ \delta \boldsymbol{r}_{d.p.}, \boldsymbol{0}_{3\times 1} \}, & \\
\end{aligned}$}
\end{equation}
which includes a zero relative velocity pre-docking waypoint $\delta \boldsymbol{r}_{w.p.}$ at time instant $t_{N_{w.p.}} \in {\mathbb{R}}_{>0}$, $N_{w.p.} \in \mathbb{N}$, and an approach cone (a.c.) constraint with approach axis $\boldsymbol{n}_{a.c.} \in {\mathbb{R}}^3$ and aperture angle $\gamma_{a.c.} \in {\mathbb{R}}$ towards the docking port located at RTN coordinates $\delta \boldsymbol{r}_{d.p.}$. In particular, the approach cone is a second-order-cone constraint with parameter matrices defined as
\begin{equation}
\resizebox{0.6\linewidth}{!}{$\begin{aligned}
    & \boldsymbol{A}_{a.c.}(t_i) = \boldsymbol{D} \boldsymbol{\Psi}(t_{i}) \in {\mathbb{R}}^{3 \times 6} \\
    & \boldsymbol{b}_{a.c.} = - \delta \boldsymbol{r}_{d. p.} \in {\mathbb{R}}^{3} \\
    & \boldsymbol{c}^T_{a.c.}(t_i) = \boldsymbol{n}^T_{a. c.} \boldsymbol{D} \boldsymbol{\Psi}(t_{i}) / \cos(\gamma_{a. c.}) \in {\mathbb{R}}^{6} \\
    & d_{a.c.} = - \boldsymbol{n}^T_{a. c.} \delta \boldsymbol{r}_{d. p.} / \cos(\gamma_{a. c.}) \in {\mathbb{R}},
\end{aligned}$}
\end{equation}
where $\boldsymbol{D} = [\boldsymbol{I}_{3 \times 3}, \boldsymbol{0}_{3 \times 3}] \in {\mathbb{R}}^{3 \times 6}$ selects the position components in the relative Cartesian state.

\subsubsection{Problem 3: Non-Convex Rendezvous Problem}
The time discretized fuel-optimal non-convex rendezvous problem is formalized as
\begin{equation}
\label{OCP_RPOD_nCVX}
\resizebox{0.98\linewidth}{!}{$\begin{aligned}
& \underset{\Delta \boldsymbol{v}(t_i), \delta \textbf{\oe}(t_i)}{\text{minimize}} & & \sum_{i = 1}^{N} || \Delta \boldsymbol{v}(t_i) ||_2 &\\
& \text{subject to} & & \delta \textbf{\oe}(t_{i+1}) = \boldsymbol{\Phi}(\delta t, t_i) \delta \textbf{\oe}(t_i) + \boldsymbol{\Phi}(\delta t, t_i) \boldsymbol{B}(t_i) \Delta \boldsymbol{v}(t_i) & \forall i \in [1, N]\\
&&& \delta \textbf{\oe}(t_1) \equiv \textrm{initial passively safe relative orbit} & \\
&&& \delta \textbf{\oe}^T(t_i) \boldsymbol{\Psi}^T(t_i) \boldsymbol{E}^T_{koz} \boldsymbol{E}_{koz} \boldsymbol{\Psi}(t_i)\delta \textbf{\oe}(t_i) \geq 1 & \forall i \in [1, N_{w.p.}] \\
&&& \boldsymbol{\Psi}(t_{N_{w.p.}}) \delta \textbf{\oe}(t_{N_{w.p.}}) = \{ \delta \boldsymbol{r}_{w.p.}, \boldsymbol{0}_{3\times 1} \} & \\
&&& || \boldsymbol{A}_{a.c.}(t_i) \delta \textbf{\oe}(t_{i}) + \boldsymbol{b}_{a.c.} ||_2 \leq \boldsymbol{c}^T_{a.c.}(t_i) \delta \textbf{\oe}(t_{i}) + d_{a.c.} & \forall i \in [N_{w.p.}, N] \\
&&& \boldsymbol{\Psi}(t_{N+1}) \delta \textbf{\oe}(t_{N+1}) = \{ \delta \boldsymbol{r}_{d.p.}, \boldsymbol{0}_{3\times 1} \}, & \\
\end{aligned}$}
\end{equation}
where $\boldsymbol{E}_{koz} = [Diag\{1/r_{r, koz}, 1/r_{t, koz}, 1/r_{n, koz}\}, \boldsymbol{0}_{3 \times 3}] \in {\mathbb{R}}^{3 \times 6}$ sizes the keep-out-zone ellipsoid around the target, with $r_{., koz} \in {\mathbb{R}}_{>0}$ the  principal semi-axes.

\subsection{Sequential Optimization}
To solve the OCP in Eq. \eqref{OCP_RPOD_nCVX} using convex optimization tools \cite{ecos_13}, the non-convex keep-out-zone constraint has to be sequentially linearized using a SCP approach \cite{scp_2000} \cite{liu_scp_2014} \cite{malyuta_scp_2022}. In particular, at iteration $k^{th}$ of the SCP loop, the non-convex constraint is linearized around a reference trajectory $\delta \bar{\textbf{\oe}}_k$, as
\begin{equation}\label{koz_lin}
\begin{aligned}
    & \boldsymbol{a}^T_{koz}(t_i) \delta \textbf{\oe}_k(t_i) \geq b_{koz}(t_i) & \forall i \in [1, N_{w.p.}],
\end{aligned}
\end{equation}
with
\begin{equation}
\resizebox{0.85\linewidth}{!}{$\begin{aligned}
    &\boldsymbol{a}^T_{koz}(t_i) = \delta \bar{\textbf{\oe}}_k^T(t_i) \boldsymbol{\Psi}^T(t_i) \boldsymbol{E}^T_{koz} \boldsymbol{E}_{koz} \boldsymbol{\Psi}(t_i) \in {\mathbb{R}}^{6} \\
    & b_{koz}(t_i) = \sqrt{\delta \bar{\textbf{\oe}}_k^T(t_i) \boldsymbol{\Psi}^T(t_i) \boldsymbol{E}^T_{koz} \boldsymbol{E}_{koz} \boldsymbol{\Psi}(t_i) \delta\bar{\textbf{\oe}}_k(t_i)} \in {\mathbb{R}}_{>0}.
\end{aligned}$}
\end{equation}
Eq. \eqref{koz_lin} is a sufficient condition for the satisfaction of the non-convex keep-out-zone constraint \cite{morgan_2014}. Moreover, to avoid phenomena as artificial unboundedness \cite{malyuta_scp_2022}, Eq. \eqref{koz_lin} has to be complemented with the second-order-cone trust region constraint
\begin{equation}
\begin{aligned}
    &|| \delta \textbf{\oe}_k(t_i) - \delta\bar{\textbf{\oe}}_k(t_i) ||_2 \leq \zeta_{k} & \forall i \in [1, N_{w.p.}]
\end{aligned}
\end{equation}
where $\zeta_{k} \in {\mathbb{R}}_{>0}$ is the trust region radius. 
This radius around the reference trajectory has to be sequentially updated by looking at the error committed in linearizing the non-convex part of the problem (in this case just the keep-out-zone constraint). 
In particular, as suggested in \cite{malyuta_scp_2022} \cite{bonalli_2019_gusto}, if the linearization error is greater than a threshold, the trust region is shrunk to not overstep the linearized model, whereas if it is lower than a threshold, the trust region can be enlarged to allow for greater exploration around the accurate linearized model. This exploration phase is usually limited to a predefined maximum number of iterations, after which, at convergence, when the linearization error tends naturally to zero, the trust region is shrunk using an exponential update rule \cite{malyuta_scp_2022}. 
Note that, the longer the exploration phase the lesser the impact of the initial warm-starting trajectory.
In this paper, the focus is on assessing the impact of the initial warm-starting trajectory on the SCP convergence performances, and as such, we focus on SCP formulations that limit the exploration to a closer neighborhood of the initial warm-start. 
Concretely, this is achieved by implementing an exponential shrink factor \cite{morgan_2014} to the trust region since the initial iterations as   
\begin{equation}
    \zeta_{k+1} = \left(\frac{\zeta_{K}}{\zeta_{1}}\right)^\frac{1}{K} \zeta_{k},
\end{equation}
where $K \in \mathbb{N}$ is the maximum number of allowed SCP loop iterations, and $\zeta_{1}$ and $\zeta_{K}$ are the user-defined initial (maximum) and final (minimum) trust region radii. 
Lastly, denoting the optimal cost of the linearized OCP at SCP iteration $k^{th}$ as ${\mathcal{J}}_k = \sum_{i = 1}^{N} || \Delta \boldsymbol{v}_k(t_i) ||_2$, the stopping criterion implemented for the SCP is represented by the following logical condition
\begin{equation}\label{SCP_stopping}
    \left(k = K \right) \lor \left( \left( \zeta_{k} \leq \zeta_{K} \right)  \land \left( {\mathcal{J}}_{k-1} - {\mathcal{J}}_{k} < {\mathcal{J}}_{tol} \right) \right),
\end{equation}
with ${\mathcal{J}}_{tol} \in {\mathbb{R}}_{>0}$ a defined tolerance on cost convergence.

\begin{table*}[ht] 
\centering
\caption{Simulation parameters.}\label{tab:param}
\resizebox{0.99\linewidth}{!}{
\begingroup
\renewcommand*{\arraystretch}{1.25}
\begin{tabular}{ c c c c c c c c c c c }
 \hline 
 \hline
  \multicolumn{2}{c}{Target orbit} & & \multicolumn{2}{c}{OCP} & & \multicolumn{2}{c}{SCP} & & \multicolumn{2}{c}{Dataset generation} \\
  \cline{1-2} \cline{4-5} \cline{7-8} \cline{10-11}
   $a$ (km) & $R_E$ + 416 & & $N$ (-) & 100 & & $K$ (-) & 20 & & $N_d$ & 400,000 \\
   $e$ (-) & 5.58e-4 & & $w_{iss}$ (m) & 108 & & $\zeta_{1}$ (m) & 200 & & Train split (\%) & 90 \\
   $i$ (deg) & 51.64 & & $l_{iss}$ (m) & 74 & & $\zeta_{K}$ (m) & 1 & & Test split (\%) & 10 \\
   $\Omega$ (deg) & 301.04 & & $h_{iss}$ (m) & 45 & & ${\mathcal{J}}_{tol}$ (m/s) & 1e-6 & & OCP horizon (orbits) & $[1, 3]$ \\
   $\omega$ (deg) & 26.18 & & $\delta \boldsymbol{r}_{d.p.}$ (m) & $\{ 0, l_{iss}, 0 \}$ & & & & & \multicolumn{1}{c}{Initial relative orbit:} &  \\
   $M (t_1)$ (deg) & 68.23 & & $\boldsymbol{n}_{a.c.}$ (m) & $\{ 0, 1, 0 \}$ & & & & & $a \delta a$ (m) & $[-5 , 5]$ \\
   period (h) & 1.548 & & $\gamma_{a.c.}$ (deg) & 30 & & & & & $a \delta \lambda$ (m) & $[-100, 100]$ \\
   & & & $\delta \boldsymbol{r}_{w.p.}$ (m) & $\delta \boldsymbol{r}_{d.p.} + 30 \boldsymbol{n}_{a.c.}$ & & & & & $a \delta e$ (m) & $r_{r, koz} + [5, 30]$ \\
   & & & $N_{w.p.}$ (-) & $N - 10$ & & & & & $a \delta i$ (m) & $r_{n, koz} + [5, 30]$ \\
   & & & $r_{r, koz}$ (m) & $h_{iss} + 15$ & & & & & $\varphi$ (deg) & $90 + [-5, 5]$ \\
   & & & $r_{t, koz}$ (m) & $l_{iss} + 20$ & & & & & $\phi$ (deg) & $90 + [-5, 5]$ \\
   & & & $r_{n, koz}$ (m) & $w_{iss} + 15$ & & & & & & \\
  \hline
  \hline
\end{tabular}
\endgroup}
\end{table*}

\begin{table*}[ht]
\centering
\caption{Forecasting Performance on Two Point Boundary Value Problem Dataset}
\small
\begingroup
\renewcommand*{\arraystretch}{1.25}
\begin{tabular}{p{3cm} l l c c c c c c c}
    \hline 
    \hline
     & & & \multicolumn{3}{c}{Transformer-only} & \multicolumn{3}{c}{Dynamics-in-the-loop} & True \\
     & Metric & & GRU & LSTM & ART & GRU & LSTM & ART & \\
    \hline
    \multirow{3}{4em}{Trajectory (RMSE)} & RTN Pos. & [m] & 1221.50 & 185.95 & 140.83 & 33.14 & 37.38 & 15.99 & - \\
     & RTN Vel. & [mm/s] & 350.33 & 128.89 & 134.39 & 16.12 & 15.60 & 10.23 & -\\
    & ROE & [m] & 880.77 & 122.73 & 85.69 & 21.49 & 24.84 & 14.95 & - \\
    \hline
    \multirow{3}{4em}{Target (RMSE)} & RTN Pos. & [m] & 3473.70 & 334.33 & 202.05 & 17.69 & 30.29 & 19.43 & -\\
    & RTN Vel. & [mm/s] & 730.30 & 190.08 & 198.67 & 11.49 & 13.18 & 6.90 & - \\
    & ROE & [m] & 2466.42 & 225.08 & 120.33 & 15.47 & 27.34 & 16.10 & - \\
    \hline
    Cost & $\sum || \Delta \boldsymbol{v} ||_2$ & [mm/s] & 396.69 & 261.52 & 241.00 & 267.66 & 260.79 & 226.14 & 220.25 \\
    Impulse Accuracy & & [\%] & 22.0 & 14.4 & 10.2 & 82.8 & 86.0 & 86.0 & - \\
    Control Error (RMSE) & & [mm/s] & 45.6& 43.2 & 38.8 & 25.7 & 22.5 & 13.6 & - \\
    Comp. Time & & [s] & 0.61 & 0.71 & 0.90 & 0.35 & 0.41 & 0.44 & 0.28 \\
    \hline
    \hline
    \end{tabular}%
    \label{tab:tpbvp}
    \endgroup
\end{table*}

\subsection{Autonomous Rendezvous MDP}
After having introduced three optimal control problems for the considered rendezvous scenario (Problems 1-3), together with a sequential optimization formulation to solve the non-convex problem (Problem 3), this section focuses on explicitly defining the elements of the MDP formulation that enables the Transformer training.
Specifically, the MDP formulation presented in Section \ref{sec_III} can be directly applied to the considered rendezvous scenario. 
The elements describing the MDP for the case of autonomous rendezvous are defined as follows:

\textit{State Space (${\calX}$):} can be expressed using either the relative Cartesian or the ROE state $\boldsymbol{x}(t_i) :\{ \delta\boldsymbol{\chi}(t_i) \lor \delta\textbf{\oe}(t_i) \} \in {\mathbb{R}}^{6}$.

\textit{Action Space (${\calU}$):} the applied delta-velocity $\boldsymbol{u}(t_i) : \Delta \boldsymbol{v}(t_i) \in {\mathbb{R}}^{3}$.

\textit{Dynamics ($f$):} the time-discretize dynamics transition is presented in Eq. \eqref{dynamics} \cite{koenig_2017_new}-\cite{chernick_2018_closed}.

\textit{Initial State Distribution ($\rho$):} the initial state distribution represents a set of passively safe relative orbits around the target outside the keep-out-zone ellipsoid, designed using relative eccentricity/inclination vectors separation \cite{damico_proximity_2006}.

\textit{Reward (${\mathcal{R}}$):} in line with the OCP formulations defined in Section \ref{sec_IV}, the reward function is defined as the negative cost function, defined as the L2-norm of the control ${\mathcal{R}}(\bx(t), \bu(t)) = -|| \Delta \boldsymbol{v}(t) ||_2$. Equivalently, the reward-to-go is formulated using Eq. \eqref{rtg} and the cost function definition, as
\begin{equation}\label{rtg_rpod}
    R(t_i) = -\sum_{j = i}^{N} || \Delta \boldsymbol{v}(t_j) ||_2
\end{equation}

\textit{Constraint-to-go ($C$):} the constraint-to-go is formulated using Eq. \eqref{ctg}, where ${\textsf{C}}(t_j)$ is defined as
\begin{equation}\label{ctg_rpod2}
\resizebox{0.89\linewidth}{!}{$
    {\textsf{C}}(t_j) = 
\begin{cases}
    1, & \text{if } \delta \textbf{\oe}^T(t_j) \boldsymbol{\Psi}^T(t_j) \boldsymbol{E}^T_{koz} \boldsymbol{E}_{koz} \boldsymbol{\Psi}(t_j)\delta \textbf{\oe}(t_j) < 1 \\
    0, & \text{otherwise}
\end{cases}$}
\end{equation}
and checks, by design, for non-convex constraint violation.

\section{Numerical Results}\label{sec_V}

In this section, we consider a rendezvous scenario of a servicer with the International Space Station. 
The orbit, OCPs, and SCP constant parameters are reported in Table \ref{tab:param} (first three columns). 
In order to generate a dataset of OCP solutions amenable for Transformer training, both the relative orbit initial conditions and the OCP control horizon are randomized within the domains specified in Table \ref{tab:param} (last column). 
Note that, while randomized, the initial relative orbital condition is enforced to be passively safe according to relative eccentricity and inclination vector separation \cite{damico_proximity_2006}.
The total dataset is comprised of 400,000 samples, which we partition according to a $90$-$10\%$ train-test split, i.e., ART is trained on $90\%$ of the data, while the remaining $10\%$ is used to generate the analyses in the remainder of this work. 
Dataset trajectories contain solutions of the non-convex problem (Problem 3) as well as solutions of the most immediate convex relaxation (Problem 2). 
For dataset generation purposes such convex relaxation is used as a warm-start for the solution of the non-convex problem.
The reader is referred to the Appendix for the specification of ART's hyperparameters. 

\subsection{Part I: Forecasting}
In this first experiment, we assess ART's performance from a forecasting perspective.
Specifically, the goal of this section is to answer the following questions: (1) are Transformers capable of imitating fuel-optimal trajectories in terms of control profile and predicted states, over long planning horizons?, (2) can we leverage knowledge of the dynamics to improve the performance of black-box Transformer networks?, and (3) computationally, what is the cost of ART compared to both optimization and ML-based approaches for the purpose of trajectory generation?

Results in Tables \ref{tab:tpbvp} and \ref{tab:scp} measure the forecasting error between the "True" fuel-optimal trajectory (i.e., trajectories obtained by solving either Problem 1 or Problem 3, respectively) and the one predicted by different learning-based methods, across different metrics.
Experiments compare ART with Gated Recurrent Unit (GRU) networks and Long Short-Term Memory (LSTM) networks: two popular recurrent neural network (RNN) architectures.
In addition to different ML architectures, this experiment also benchmarks the performance of the two trajectory generation strategies introduced in Algorithms \ref{Algo_inf_tonly} and \ref{Algo_inf_dyn}.
In both experiments, forecasting performance is evaluated with respect to (i) state deviation along the entire trajectory, in both RTN and ROE representations, (ii) target (i.e., terminal) state deviation, in both RTN and ROE representations, (iii) overall cost and deviation of the predicted control profile, and (iv) computation time to generate a full open-loop trajectory.
Specifically, Table \ref{tab:tpbvp} shows results evaluated on the Two Point Boundary Value Problem (Section \ref{sec_IV}, Problem 1), while Table \ref{tab:scp} focuses on the Non-Convex Rendezvous Problem (Section \ref{sec_IV}, Problem 3), thus quantifying prediction performance on imitating trajectories computed through the solution of both a convex and non-convex problem, respectively.
Results measure the following performance metrics averaged over the entire test set (i.e., $\approx20,000$ trajectories):
\begin{enumerate}
    \item \textit{Trajectory (RMSE)}: root-mean-squared-error evaluated over the states of the entire trajectory.
    \item \textit{Target (RMSE)}: root-mean-squared-error evaluated between the target condition and the final state in the predicted trajectory.
    \item \textit{Cost}: total cost obtained by the trajectory.
    \item \textit{Impulse accuracy}: measures the temporal accuracy of the predicted controls with respect to the control profile of the ground-truth trajectory. Concretely, an accuracy of $100\%$ represents a situation for which whenever "true" applies a $\Delta \boldsymbol{v}(t_i) > 0$ at time instant $t_i$, the predicted trajectory also applies $\Delta \boldsymbol{v}(t_i) > 0$ at time instant $t_i$. 
    \item \textit{Control Error (RMSE)}: root-mean-squared-error between "true" and predicted $\Delta \boldsymbol{v}(t_i)$.
    \item \textit{Computation Time}: runtime to generate a full trajectory.
\end{enumerate}

\begin{table*}[ht]
\centering
\caption{Forecasting Performance on Non-Convex Rendezvous Problem Dataset}
\small
\begingroup
\renewcommand*{\arraystretch}{1.25}
\begin{tabular}{p{3cm} l l c c c c c c c}
    \hline 
    \hline 
     & & & \multicolumn{3}{c}{Transformer-only} & \multicolumn{3}{c}{Dynamics-in-the-loop} & True \\
     & Metric & & GRU & LSTM & ART & GRU & LSTM & ART & \\
    \hline 
    \multirow{3}{4em}{Trajectory (RMSE)} & RTN Pos. & [m] & 82.19 & 31.05 & 25.02 & 18.62 & 17.69 & 17.94 & - \\
     & RTN Vel. & [mm/s] & 116.24 & 17.30 & 15.08 & 14.01 & 13.03 & 12.35 & -\\
    & ROE & [m] & 48.28 & 22.62 & 17.17 & 14.74 & 13.42 & 13.18 & - \\
    \hline
    \multirow{3}{4em}{Target (RMSE)} & RTN Pos. & [m] & 280.04 & 59.52 & 34.90 & 19.59  & 16.59 & 16.83 & -\\
    & RTN Vel. & [mm/s] & 83.52 & 26.75 & 14.81 & 31.18 & 27.82 & 20.65 & - \\
    & ROE & [m] & 117.76 & 46.34 & 25.31 & 34.33 & 29.32 & 25.56 & - \\
    \hline
    Cost & $\sum || \Delta \boldsymbol{v} ||_2$ & [mm/s] & 362.42 & 254.76 & 249.62 & 294.95 & 284.37 & 257.28 & 241.34 \\
    Comp. Time & & [s] & 0.61 & 0.71 & 0.90 & 0.35 & 0.41 & 0.44 & 2.21 \\
    \hline
    \hline
    \end{tabular}%
    \label{tab:scp}
    \endgroup
\end{table*}

Results in both tables show that ART is able to consistently outperform other learning-based approaches across all performance metrics and across both inference strategies (i.e., "transformer-only" and "dynamics-in-the-loop").
Interestingly, the results show a substantial benefit in leveraging knowledge of the dynamics, with predictions computed under dynamics-in-the-loop inference clearly outperforming the ones computed using Transformers as an approximate dynamics model. 
Crucially, the ability to propagate states and achieve dynamically feasible trajectories substantially improves the ability of Transformers to imitate optimal control profiles, as observable in the \textit{Impulse accuracy} and \textit{Control Error} metrics, with an improvement of $\approx 60\%$ in accuracy, on average (Table \ref{tab:tpbvp}).
Furthermore, consistently with ART's accurate control profile, the proposed ART-dynamics-in-the-loop strategy is able to substantially outperform other approaches with respect to the overall \textit{Cost}, even in those cases where the pure state prediction performance is comparable to alternatives, as in Table \ref{tab:scp}.
Computationally, results show how all learning-based approaches are typically more expensive when compared to the solution of a convex program (Table \ref{tab:tpbvp}). 
This order is reversed for the non-convex case, with ART trajectory generation being approximately five times faster compared to the solution of the SCP (Table \ref{tab:scp}).
Results also show how the use of the dynamics achieves faster trajectory generation, thus further highlighting the benefits of the inference strategy depicted in Algorithm \ref{Algo_inf_dyn}.


\subsection{Predicting Future Performance Parameters}
\begin{figure}
\centering
   \includegraphics[width=0.95\columnwidth]{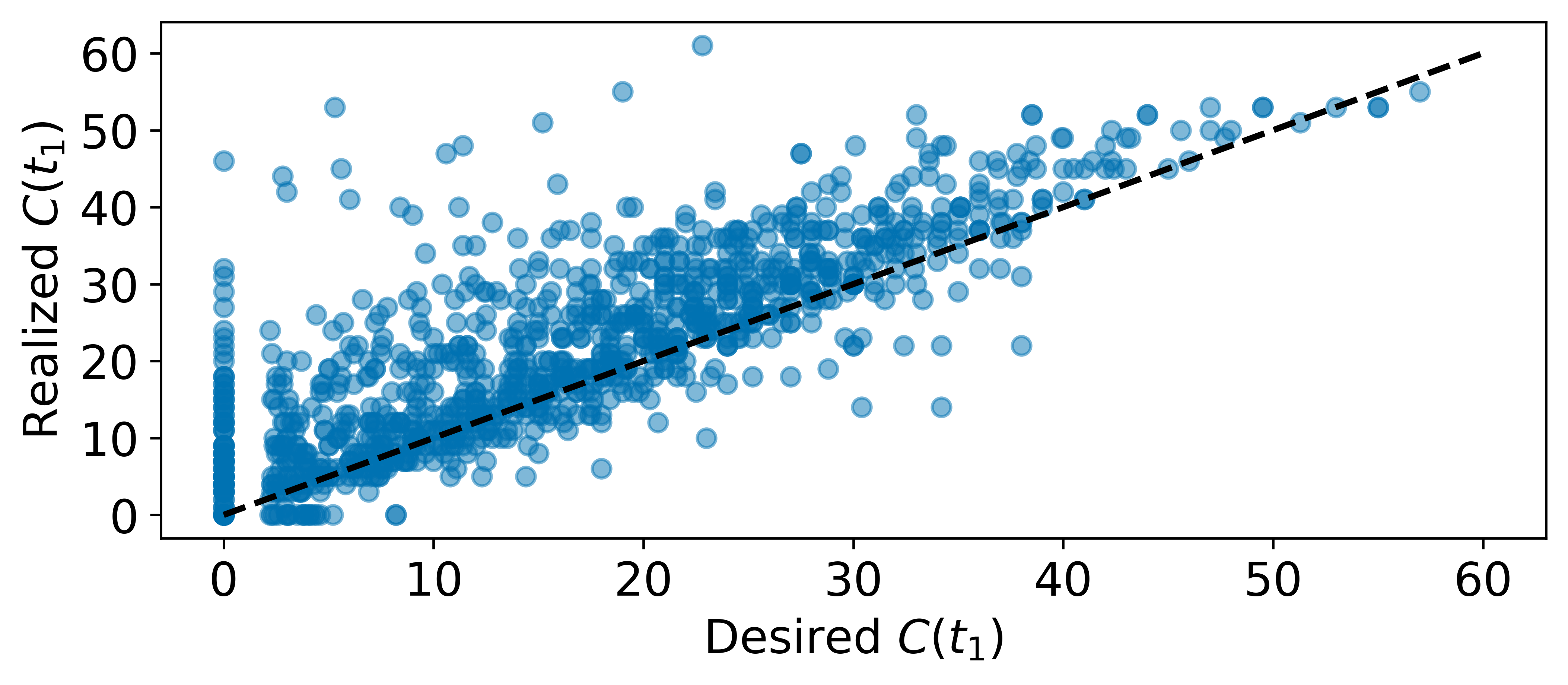}
   \label{fig:Ng1} 
   \includegraphics[width=0.95\columnwidth]{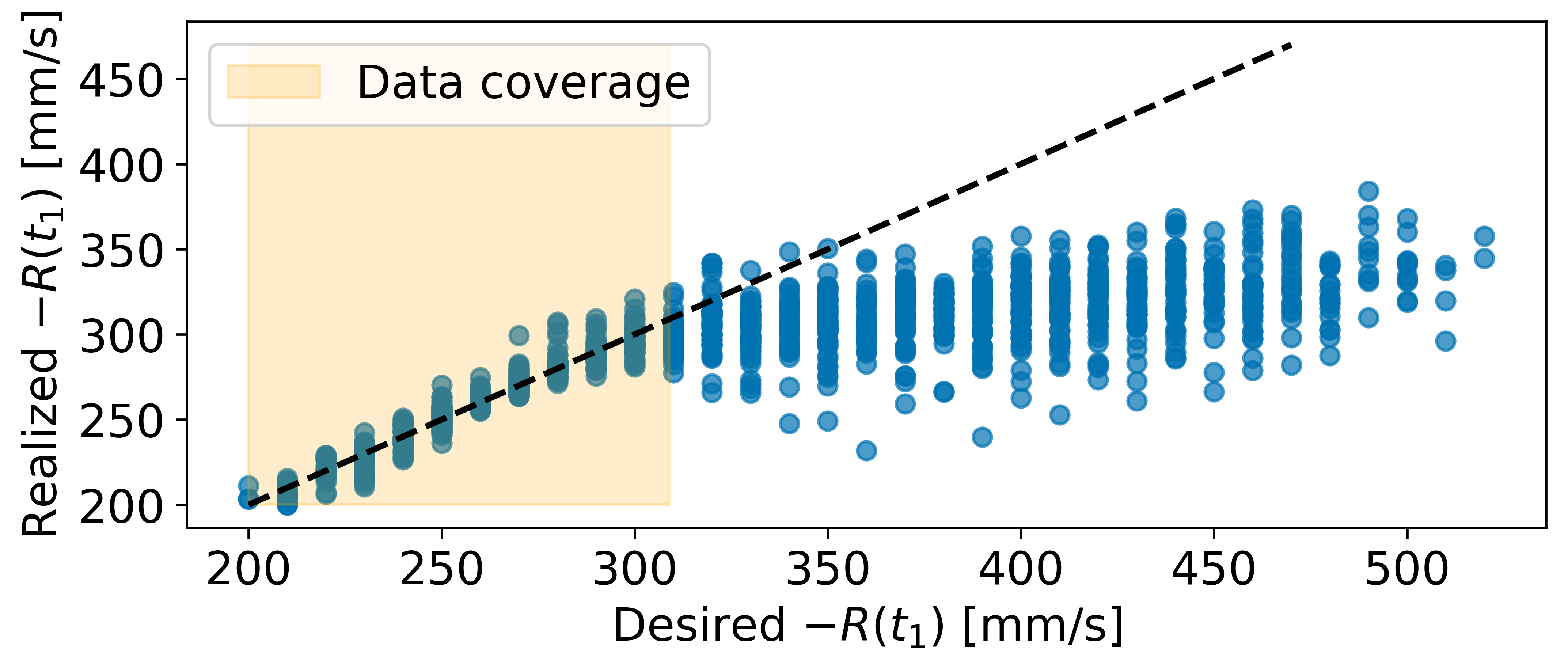}
   \label{fig:Ng2}
   \caption{Constraint violations (top) and costs (bottom) realized by ART-predicted trajectories when conditioned on desired (user-specified) performance parameters.}
   \label{fig:ctg_rtg_analysis}
\end{figure}
As a further experiment, Figure \ref{fig:ctg_rtg_analysis} evaluates the ability of ART to model trajectory performance parameters. 
The results show the constraint violation (top), and cost (bottom, expressed as negative reward-to-go) achieved by the trajectories predicted by ART for varying values of desired constraint-to-go and reward-to-go, respectively.
Specifically, Figure \ref{fig:ctg_rtg_analysis} shows the alignment between the user-specified performance parameter (x-axis) and the actual value achieved when executing the respective trajectory predicted by ART (y-axis), on test data.
In both analyses, the desired and realized performance parameters are highly correlated, especially if the desired parameter lies in the range of values available in the dataset (i.e., the entire x-axis for the constraint-to-go and the yellow-shaded region for the reward-to-go).
As a result, these plots highlight how, once trained, ART is able to replicate specific configurations of the performance parameter reliably, thus enabling a novel degree of control over the output of learning-based components. 
Finally, looking at the discrepancy between the blue dots and the ideal trend (black dashed line) in Figure \ref{fig:ctg_rtg_analysis} on the top, it is relevant to emphasize that ART trajectory generation does not provide hard guarantees of constraint satisfaction. Such guarantees can be retrieved by using the trajectory generated by ART as a warm-start for a sequential convex program.

\subsection{Part II: Control}

\begin{figure*}[ht]
    \centering
    \includegraphics[width=0.49\linewidth]{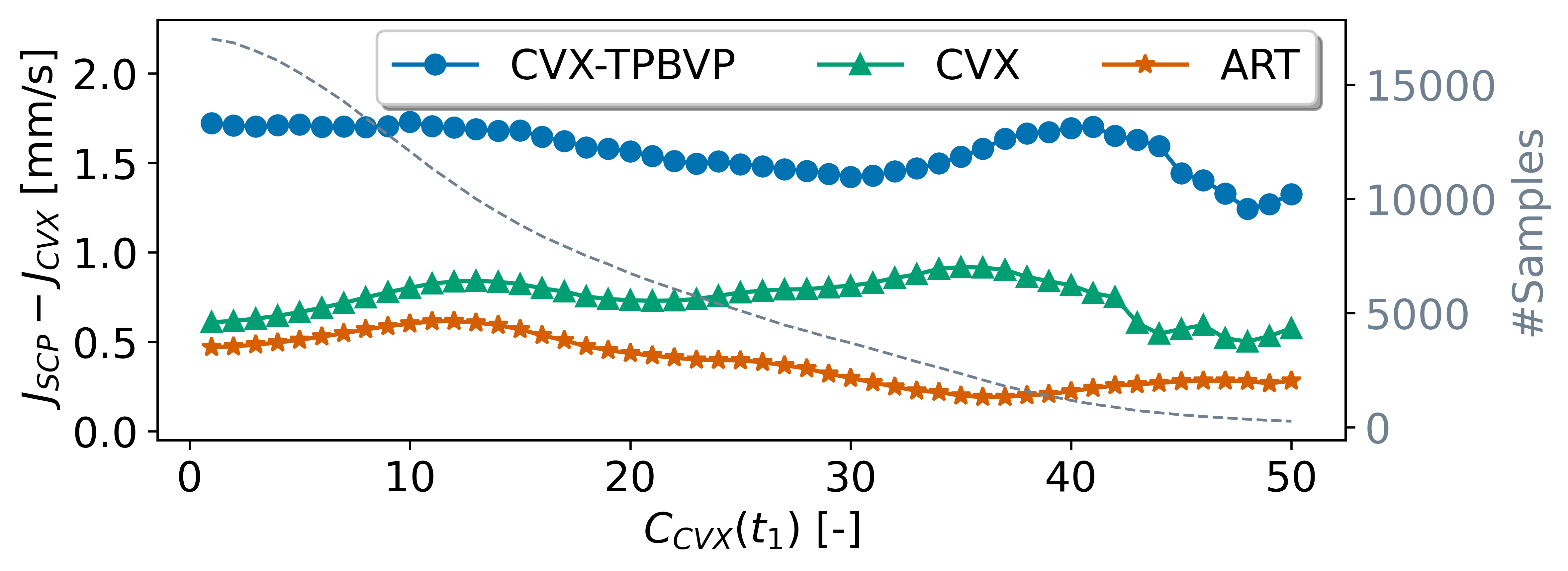}\hspace*{0.1cm}\includegraphics[width=0.49\linewidth]{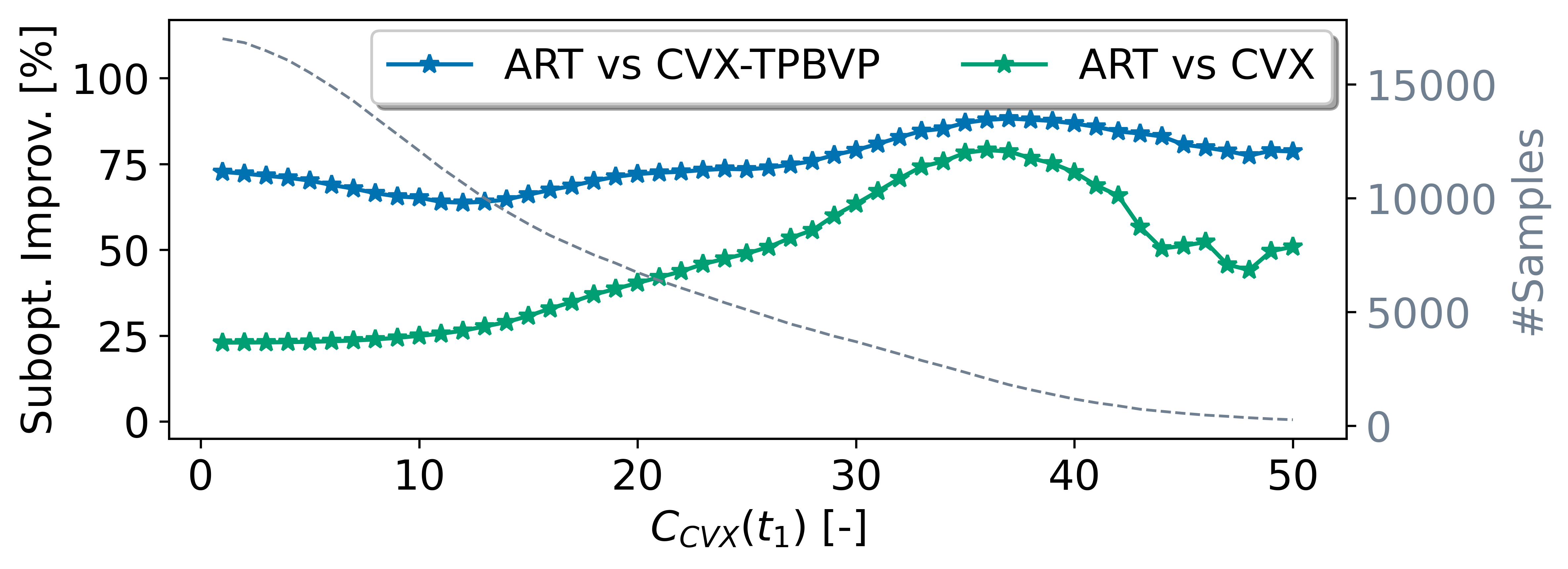}
     \caption{Control cost comparison. (Left) Absolute sub-optimality gap of different SCP solutions with respect to the convex lower bound. (Right) Percentage improvement brought by ART with respect to the two convex relaxations considered in this work.} \label{fig:cost_ws}
\end{figure*}

\begin{figure*}[ht]
    \centering
    \includegraphics[width=0.49\linewidth]{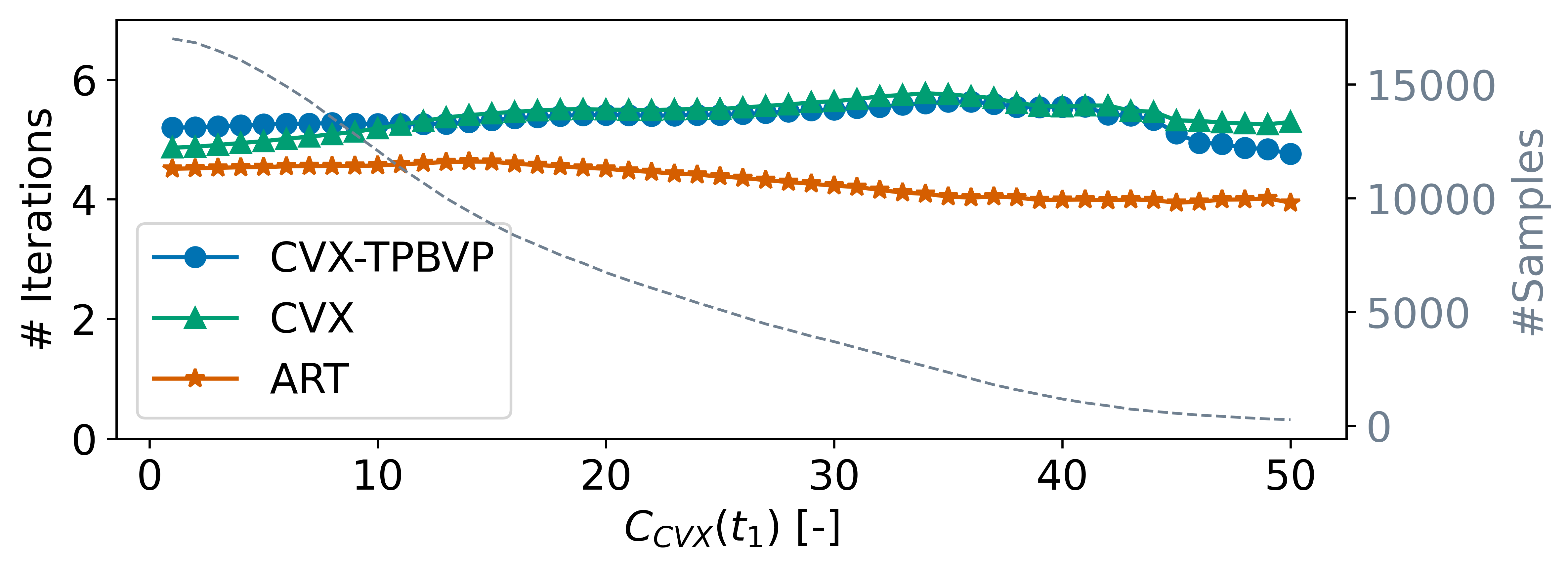}\hspace*{0.1cm}\includegraphics[width=0.47\linewidth]{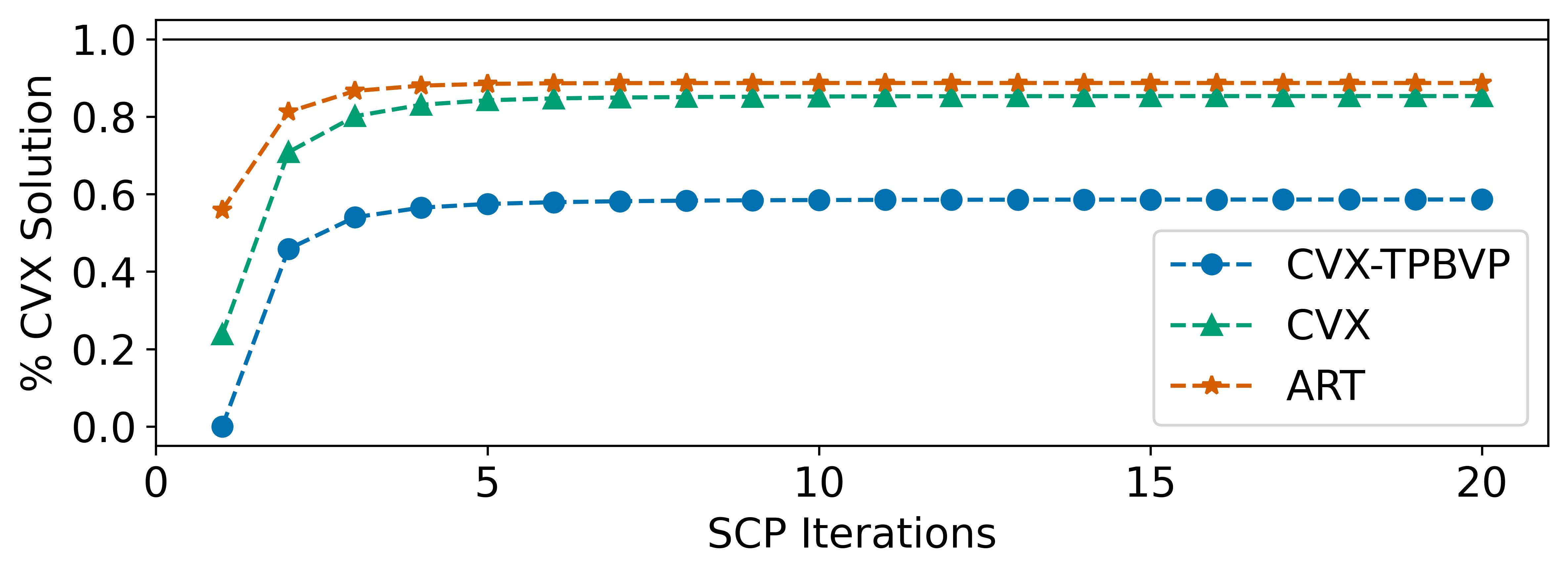}
     \caption{SCP iterations comparison. (Left) The number of iterations required to obtain SCP convergence. (Right) Cost convergence across SCP iterations.} 
     \label{fig:iter_ws}
\end{figure*}

\begin{figure*}[ht]
    \centering
    \includegraphics[width=0.49\linewidth]{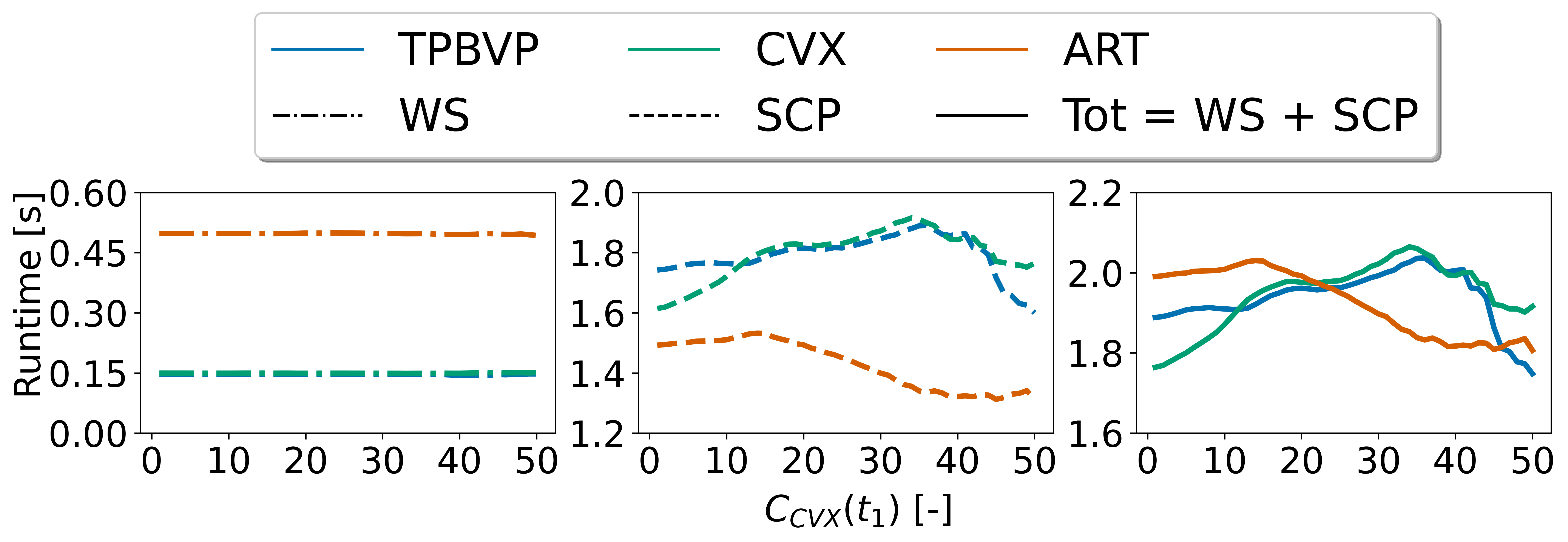}\hspace*{0.1cm}\includegraphics[width=0.47\linewidth]{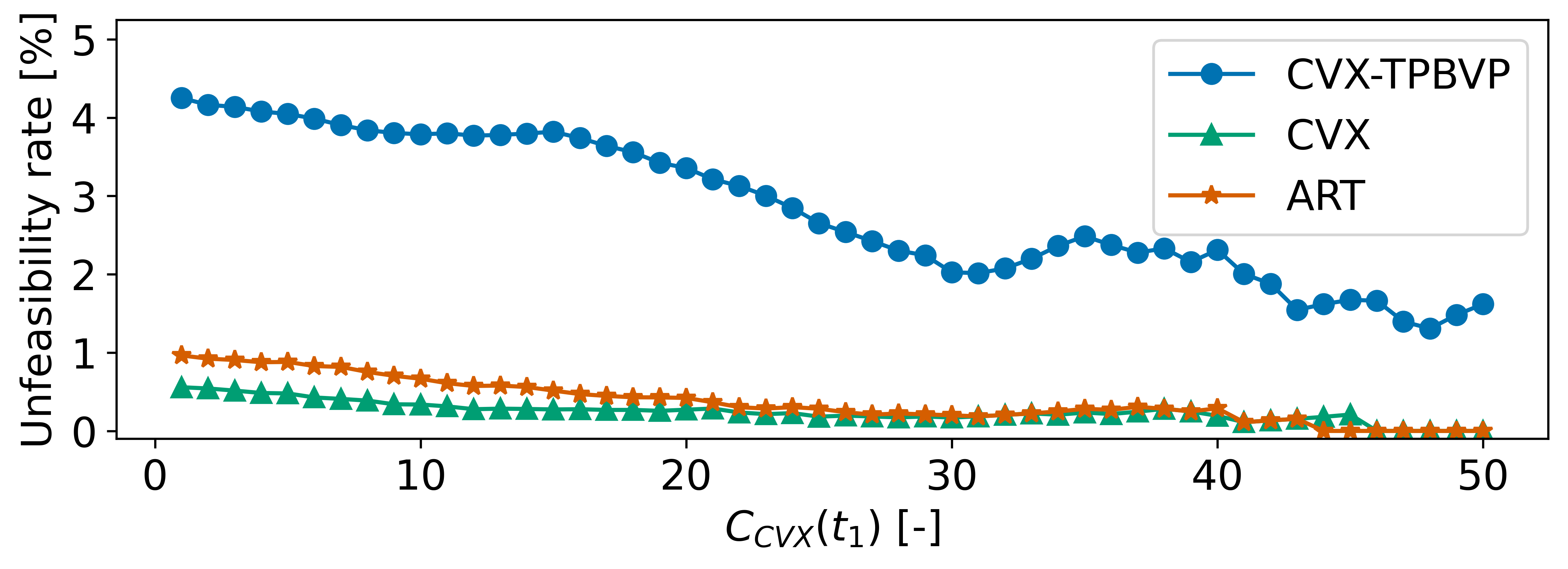}
     \caption{(Left) Runtime comparison. (Right) SCP unfeasibility rate.} 
     \label{fig:runtime_unfeas_ws}
\end{figure*}

After having assessed ART's performance from a forecasting standpoint, this section focuses on analyzing the ability of Transformers to generate effective SCP warm-starts for solving the non-convex rendezvous OCP (Problem 3, in Section \ref{sec_IV}).
In particular, as a form of comparison, we warm-start the SCP using: (i) a trajectory obtained solving the convex two-point-boundary-value-problem from the initial condition to the docking port (Problem 1, labeled as "CVX-TPBVP"), (ii) a trajectory obtained solving the convex rendezvous problem which does not enforce the keep-out-zone constraint (Problem 2, labeled as "CVX"), and (iii) a trajectory generated by ART using Algorithm \ref{Algo_inf_dyn}.
As discussed in Section \ref{sec_III}, to initialize Algorithm \ref{Algo_inf_dyn} we set the performance parameter $R(t_1)$ to the negative total cost of the "CVX" solution and $C(t_1) = 0$.

Figures \ref{fig:cost_ws}-\ref{fig:runtime_unfeas_ws} assess the following performance metrics: (i) the achieved optimality level, (ii) the number of iterations required for the SCP to converge (given the stopping criteria defined in Eq. \eqref{SCP_stopping}), (iii) the associated runtime (broken down in time needed to compute the warm-start and time needed for running the SCP), and (iv) the unfeasibility rate of the SCP (where a given problem instance is deemed to be unfeasible if the convex solver \cite{ecos_13} is unable to find any feasible optimal solution through all SCP iterations). 
These metrics are presented as a function of the initial constraint-to-go observed for the convex rendezvous problem solution $C_{CVX}(t_1)$, which represents a direct metric of the difficulty of the considered SCP warm-starting scenario.
Specifically, scenarios where $C_{CVX}(t_1) = 0$ represent OCPs for which a convex solution non-violating the keep-out-zone constraint already exists, and the associated global optimum is given by the cost of this convex solution.
Whereas, scenarios where $C_{CVX}(t_1) >> 0$ represent OCPs for which the closest convex relaxation is highly unfeasible and where the selection of a higher-quality initial warm-start can have a greater impact on SCP performance.
Within the considered test dataset, approximately half of the trajectories correspond to a scenario for which $C_{CVX}(t_1) > 0$. In Figures \ref{fig:cost_ws}-\ref{fig:runtime_unfeas_ws}, the grey dashed line presents the amount of test data samples for which $C_{CVX}(t_1)$ is greater than the value reported on the x-axis.
The displayed performance metrics are average values over the number of test samples indicated by this grey dashed line.

Figure \ref{fig:cost_ws} presents the optimality metric results. On the left, the sub-optimality gap of the SCP solutions with respect to the lower-bounding global optimum of the convex rendezvous problem is presented for the three considered warm-starting cases. On the right, the corresponding percentage improvement brought by using ART's trajectory as a warm-start is presented. 
In particular, ART warm-start introduces an average improvement of more than $20\%$ for all $C_{CVX}(t_1)>0$ values, meaning that, on average, Transformer-generated warm-starts lead the SCP to converge to more fuel-efficient trajectories.
Moreover, looking at the range $C_{CVX}(t_1) \in [30, 40]$, the percentage improvement brought by ART is up to $80\%$ with respect to the "CVX" warm-start, and up to $\approx 90\%$ with respect to the "CVX-TPBVP" warm-start. 
This $C_{CVX}(t_1)$ range represents scenarios for which the keep-out-zone constraint was violated for up to $40\%$ of the time steps by the "CVX" solution. 
This shows how ART is capable of outperforming the convex benchmarks, especially in the most challenging warm-starting scenarios, where even the most immediate convex relaxation is far from being a feasible solution for the non-convex OCP.
Note that the decreasing behavior of ART performance for $C_{CVX}(t_1) > 40$ may be attributed to the lower number of test samples available for computing the average performance metric.

As a further analysis, Figure \ref{fig:iter_ws} focuses on the iterations required for SCP convergence. 
On the left, the average number of required iterations is presented for the three considered warm-starts. 
Results show how ART-generated warm-starts require on average a lower number of iterations for SCP convergence with respect to the convex benchmarks. 
This up to $\approx 1.5$ iterations in the range $C_{CVX}(t_1) \in [30, 40]$.
On the right, the convergence of the cost throughout the SCP iterations is analyzed, and presented as a percentage achievement of the lower-bounding global optimum of the convex rendezvous problem (for $C_{CVX}(t_1)>1$). 
In particular, consistently with the results presented in Figure \ref{fig:cost_ws} and Figure \ref{fig:iter_ws} on the left, ART warm-start fosters convergence to a more fuel-efficient trajectory in fewer SCP iterations.

\begin{figure*}[t]
    \centering
   \includegraphics[align=c, width=0.67\linewidth]{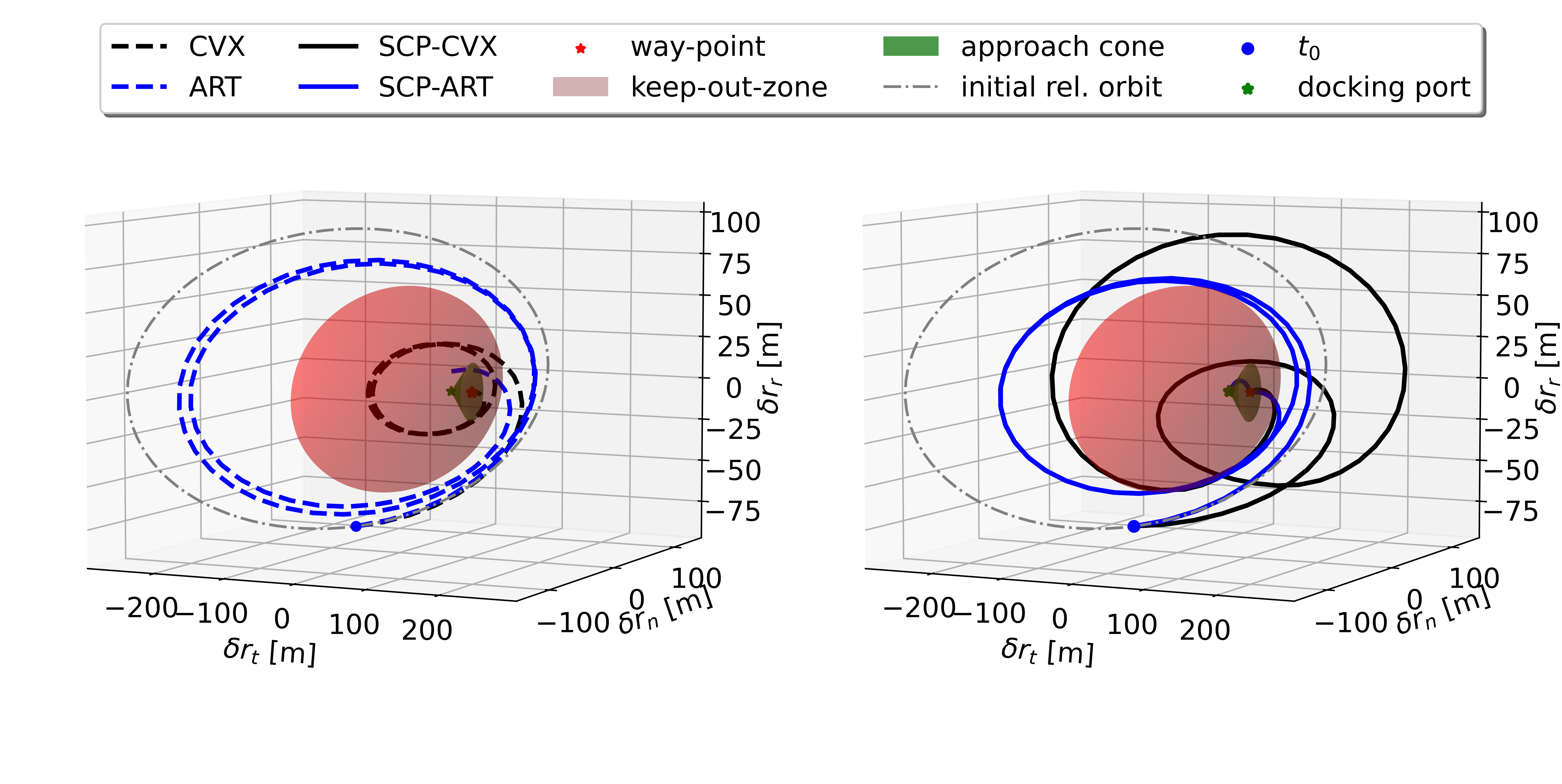}\hspace*{0.4cm}\includegraphics[align=c, width=0.3\linewidth]{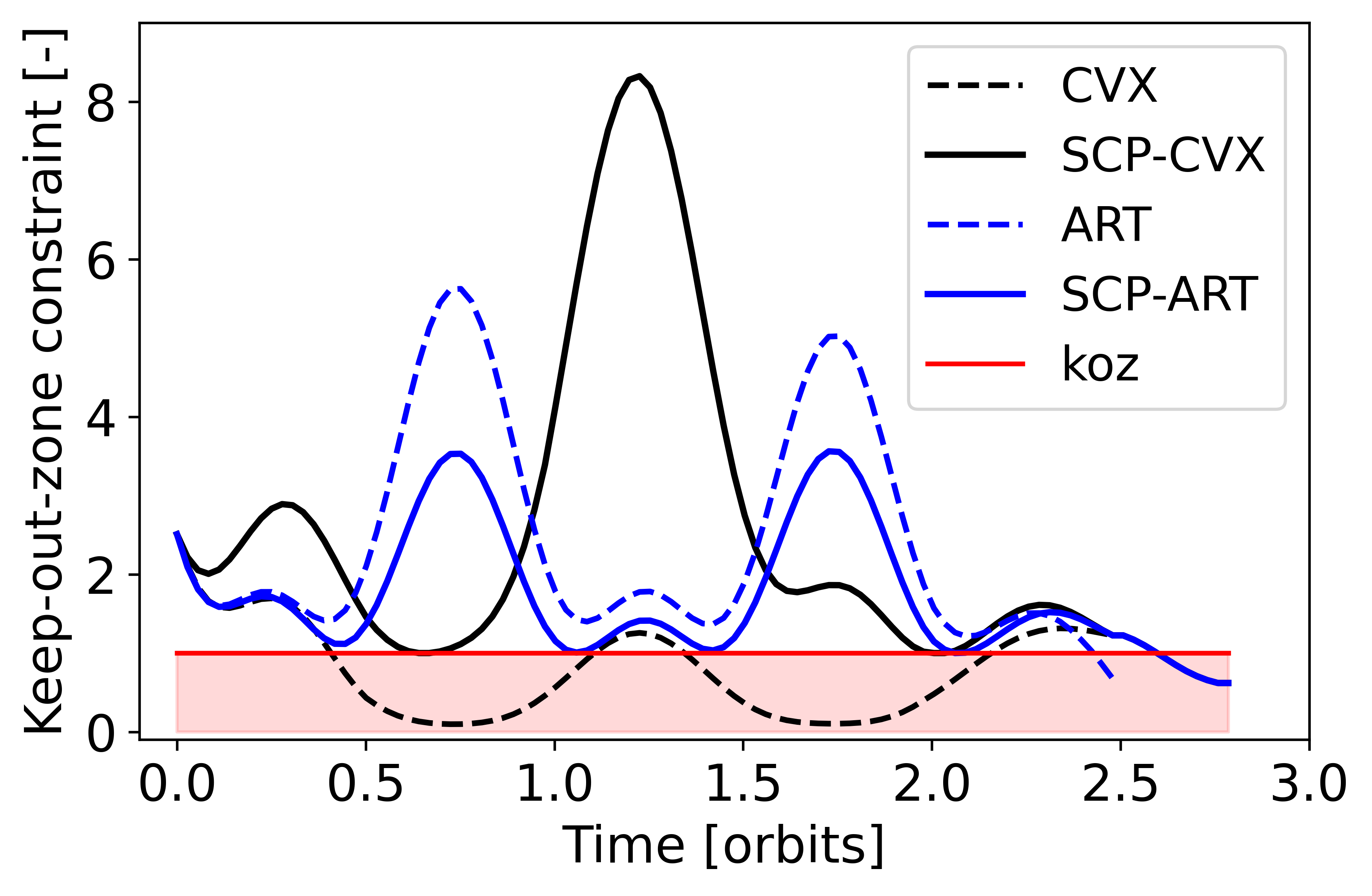}
     \caption{Cartesian RTN trajectories (left), keep-out-zone costraint satisfaction (right).} \label{fig:cart_traj}
\end{figure*}

\begin{figure*}[t]
    \centering
   \includegraphics[width=0.49\linewidth]{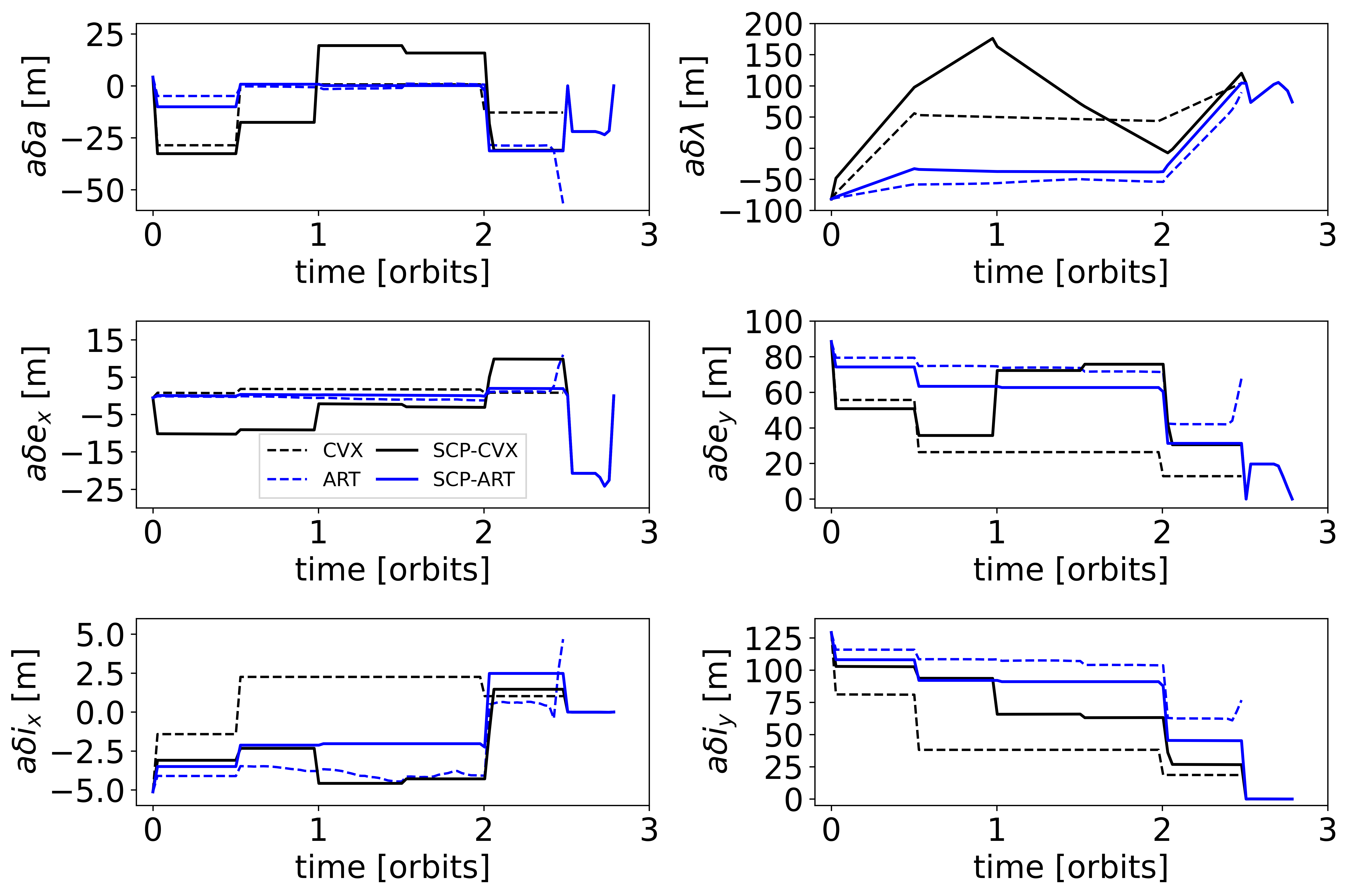}\hspace*{0.1cm} \includegraphics[width=0.49\linewidth]{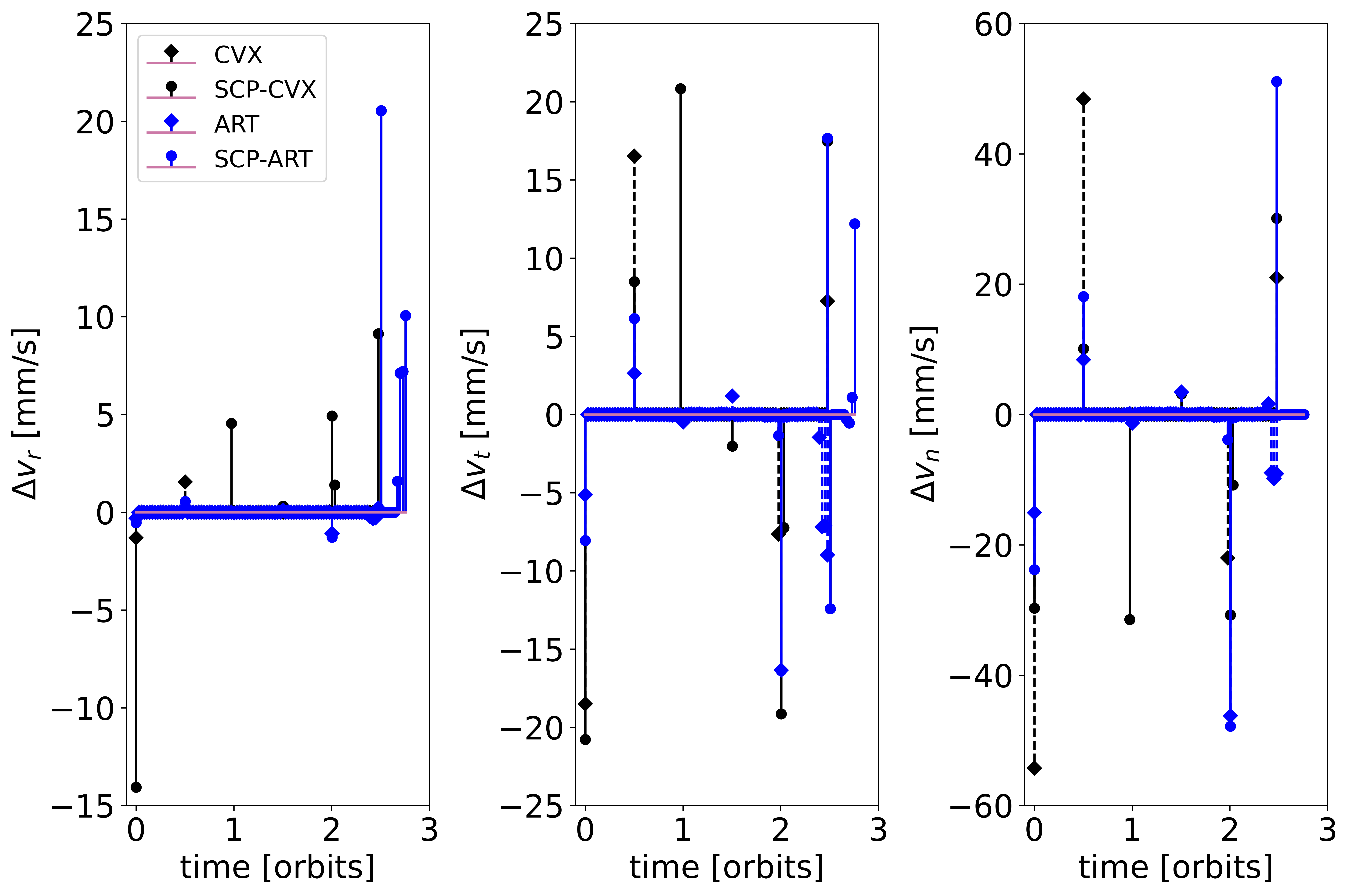}
     \caption{ROE trajectories (left), maneuver profiles (right).} \label{fig:roe_man}
\end{figure*}

To further assess the computational cost of different methods, Figure \ref{fig:runtime_unfeas_ws} (on the left) analyzes the average required runtime\footnote{Computation times are from a Linux system equipped with a 3.60GHz processor and 32GB RAM, and a NVIDIA RTX 2080 Ti 12GB GPU.} broken down in (i) time needed to compute the warm-start (dot-dashed lines, "WS"), (ii) time needed for running the SCP (dashed-lines, "SCP"), and (iii) total time (continuous lines, "Tot = WS + SCP"), for the three considered warm-starts. 
ART trajectory inference (Algorithm \ref{Algo_inf_dyn}) takes an average $\approx0.5s$, whereas solving the convex optimization problems takes an average $\approx0.15s$. 
The higher computational overhead in generating the warm-start using ART is expected, given the high computational efficiency of solving convex optimization problems via off-the-shelf solvers \cite{ecos_13}. 
Nevertheless, the gain in SCP iterations brought by ART (Figure \ref{fig:iter_ws}) is reflected in the runtime required for solving the SCP, which is up to $1s$ faster for $C_{CVX}(t_1) \in [30, 40]$. This compensates for the higher runtime needed to compute the warm-start, thus leading to analogous total runtimes of $\approx2s$ for the three considered warm-starts, with an advantage brought by ART of up to $\approx0.3s$ for high $C_{CVX}(t_1)$. 
To further reduce the runtime of ART, we believe future work could explore recent advances regarding inference-time optimization for the Transformer architecture in Large Language Models (LLMs)\cite{TayEtAl2022}.



Finally, Figure \ref{fig:runtime_unfeas_ws} (on the right) presents the average unfeasibility rate of SCP convergence over the analyzed test samples for the three considered warm-starts. 
In particular, a given problem instance is deemed to be unfeasible if the convex solver \cite{ecos_13} is unable to find any feasible optimal solution through all SCP iterations.
In this case, the "CVX" warm-starting benchmark provides the lowest SCP unfeasibility rate, up to $\approx0.5\%$ better than ART, which in turn outperforms "CVX-TPBVP" by up to $\approx3 \%$. 
This result shows how the most immediate convex relaxation ("CVX") is a reliable warm-start and compelling benchmark to compare ART to. 
The fact that ART performs similarly to "CVX" while out-performing "CVX-TPBVP" can be seen as an indicator of the reliability of ART trajectory generation.


\subsection{Qualitative Assessment}

To gain a qualitative and visual understanding of the potential benefits of ART warm-starting trajectory generation, Figure \ref{fig:cart_traj} focuses on showing the trajectories for a specific test sample, one for which ART warm-start leads to a substantially different solution that is $\approx11\%$ more fuel-efficient than "CVX" warm-start, in $16$ fewer SCP iterations.
Specifically, Figure \ref{fig:cart_traj} presents the cartesian rendezvous trajectories in the RTN reference frame (left) together with the corresponding values of the keep-out-zone constraint (right). 
The warm-starts are dashed, and the corresponding SCP trajectories are depicted by continuous lines. 
By focusing on the keep-out-zone constraint, it is interesting to notice how, in the interval $[1, N_{w.p.}]$, the "CVX" warm-start is largely unfeasible, thus violating the keep-out-zone constraint multiple times. On the other hand, the ART warm-start is almost always feasible. 
Finally, Figure \ref{fig:roe_man} presents the corresponding ROE trends (left) together with the corresponding maneuver profiles (right). Looking at the ROE trends, it is interesting to observe further how the ART warm-start places the initial trajectory closer to the final SCP solution than the "CVX" warm-start, thereby representing a better initial guess for the non-convex optimization problem.

\section{Conclusions}

The use of AI for spacecraft on-board autonomy, in both theory and practice, is typically challenged by the lack of constraint satisfaction guarantees and expensive simulation, potentially hindering real-world adoption.
In this paper, we assess the capability of modern generative models to solve complex trajectory optimization problems by proposing the \textit{Autonomous Rendezvous Transformer} for spacecraft rendezvous.
Specifically, instead of approaching the problem purely end-to-end, with ML models directly mapping from states to controls, we propose the use of Transformers to warm-start a sequential convex program with the objective of (i) enhancing traditional optimization-based methods via ML, and (ii) enforcing hard constraint satisfaction within learning-based approaches via optimization.
Empirically, results from a forecasting and control standpoint show how ART is able to generate near-optimal trajectories efficiently and outperform challenging benchmarks.
This work further investigates crucial design decisions and highlights a collection of desirable features that emerge from training modern generative models within a sequential decision-making setting, such as the ability to accurately model trajectory performance parameters.
This work opens several promising research directions including (i) investigating the robustness of the proposed methodology in presence of uncertainties (e.g., stemming from navigation, actuation, and unmodeled system dynamics), and (ii) the application to more realistic closed-loop control scenarios, even in the event of contingencies. More generally, we believe this research opens several promising directions for the extension of this framework to a broader class of complex trajectory optimization problems within spacecraft applications.



\appendix{}\label{appendix:hyper} 
The Autonomous Rendezvous Transformer presented in this work is implemented in PyTorch \cite{paszke2017automatic} and builds off Huggingface's \code{transformers} library \cite{HuggFaceTransf}.
Specifically, Table \ref{tab_appendix:hyper} presents an overview of the hyperparameter settings used in this work.
\begin{table}[ht]
\centering
\caption{Hyperameters of ART architecture for the autonomous rendezvous experiments.}
\begingroup
\renewcommand*{\arraystretch}{1.25}
\begin{tabular}{l l}
    \hline 
    \hline
     Hyperparameter & Value \\
    \hline
     Number of layers & 6\\
     Number of attention heads & 6 \\
    Embedding dimension & 384 \\
     Batch size& 4 \\
    Context length $K$ & 100 \\
    Non-linearity & ReLU\\
    Dropout & 0.1\\
    Learning rate &  $3e^{-5}$\\
    Grad norm clip & 1.0 \\
    Learning rate decay & None \\
    Gradient accumulation iters & 8 \\
    \hline
    \hline
    \end{tabular}%
    \label{tab_appendix:hyper}
    \endgroup
\end{table}

\acknowledgments
The authors thank Davide Celestini for his help with the analyses and simulations. This work is supported by Blue Origin (SPO \#299266) as
Associate Member and Co-Founder of the Stanford’s Center of AEroSpace Autonomy Research (CAESAR) and the NASA University Leadership Initiative (grant \#80NSSC20M0163). This article solely reflects the opinions and conclusions of its authors and not any Blue Origin or NASA entity.

\bibliographystyle{IEEEtran}
\bibliography{main} 

\clearpage

\thebiography
\begin{biographywithpic}
{Tommaso Guffanti}{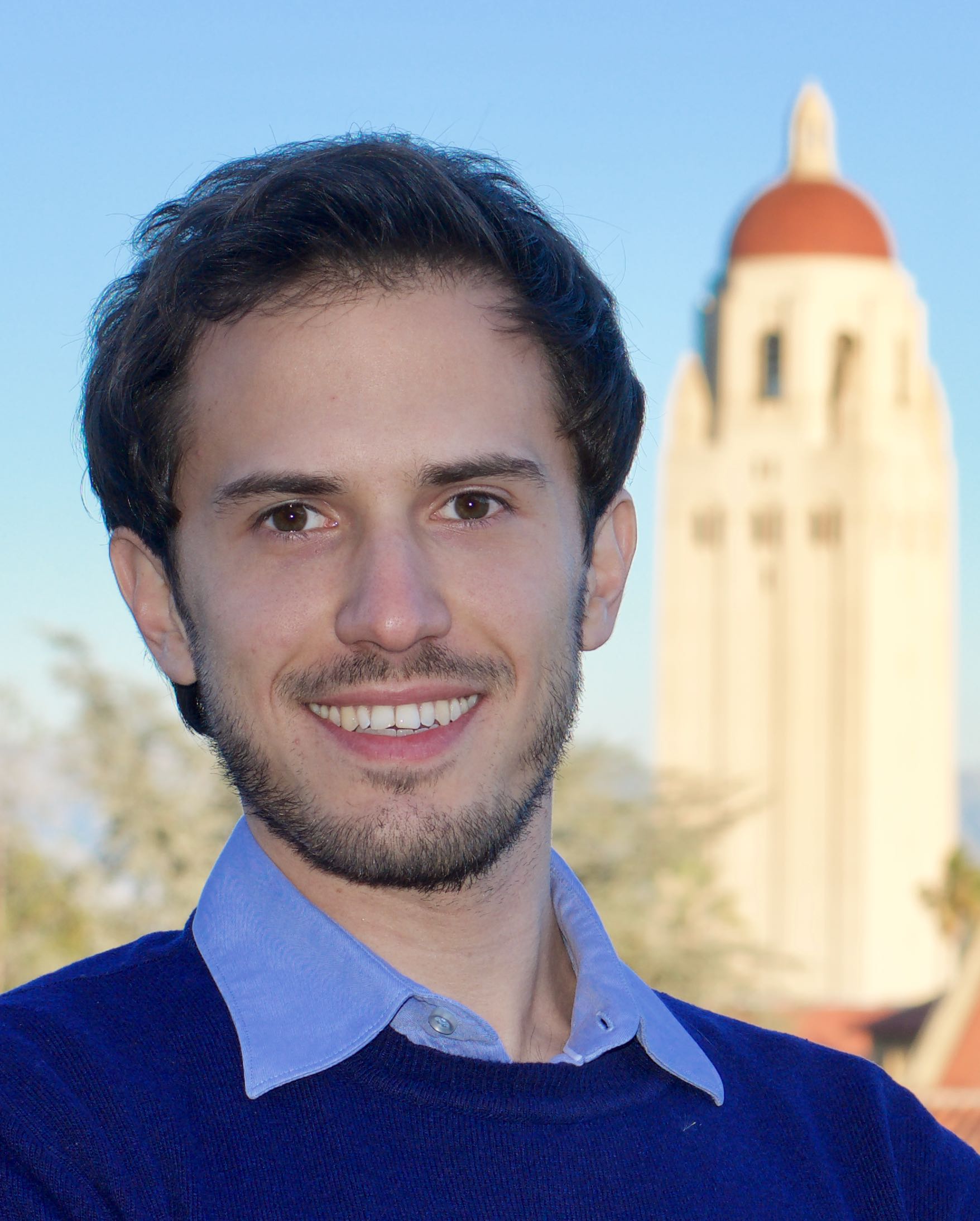}
Dr. Tommaso Guffanti is a Postdoctoral Scholar in the Space Rendezvous Lab at Stanford University. He received the B.S. and M.S. degrees in Aerospace engineering cum laude from Politecnico di Milano, and the Ph.D. degree in Aeronautics and Astronautics from Stanford University. Dr. Guffanti research contributions aim at developing cutting-edge guidance and control algorithms, and flight software to enable safe and autonomous functions and operations on-board space vehicles, in order to satisfy the requirements of the next generation of multi-spacecraft missions. During his doctorate and postdoctorate, he has been making research contributions in astrodynamics, safe and fault-tolerant multi-agent optimal control, and machine learning-based spacecraft motion planning and guidance for a variety of projects funded by national agencies and industry. He has over 15 scientific publications, including conference proceedings, and peer-reviewed journal articles. He has been awarded a doctoral Stanford Graduate Fellowship, a post-doctoral Stanford Center of Excellence for Aeronautics and Astronautics Scholarship, and a Stanford Emerging Technology Review Fellowship. He has been named excellent reviewer of the Journal of Guidance, Control, and Dynamics for three years.
\end{biographywithpic} 

\begin{biographywithpic}
{Daniele Gammelli}{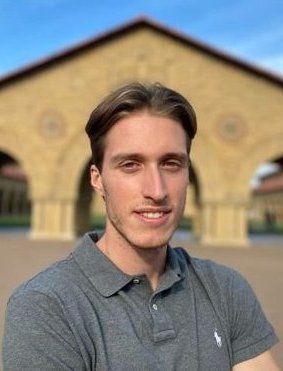}
Dr. Daniele Gammelli is a Postdoctoral Scholar in the Autonomous Systems Lab at Stanford University. 
He received the Ph.D. in Machine Learning and Mathematical Optimization at the Department of Technology, Management and Economics at the Technical University of Denmark. 
Dr. Gammelli's research focuses on developing learning-based solutions that enable the deployment of future autonomous systems in complex environments, with an emphasis on large-scale robotic networks, aerospace systems, and future mobility systems.
During his doctorate and postdoctorate career, Dr. Gammelli's has been making research contributions in fundamental AI research, robotics, and its applications to network optimization and mobility systems.
His research interests include deep reinforcement learning, generative models, graph neural networks, bayesian statistics, and control techniques leveraging these tools.
\end{biographywithpic}

\begin{biographywithpic}
{Simone D'Amico}{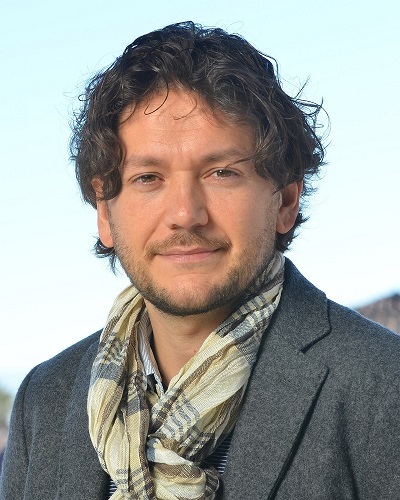}
Dr. Simone D'Amico is an Associate Professor of Aeronautics and Astronautics (AA), W.M. Keck Faculty Scholar in the School of Engineering, and Professor of Geophysics (by Courtesy). He is the Founding Director of the Stanford's Space Rendezvous Laboratory and Director of the AA Undergraduate Program. He received the B.S. and M.S. degrees from Politecnico di Milano (2003) and the Ph.D. degree from Delft University of Technology (2010). Before Stanford, Dr. D'Amico was research scientist and team leader at the German Aerospace Center (DLR) for 11 years. There he gave key contributions to formation-flying and proximity operations missions such as GRACE, PRISMA, TanDEM-X, BIROS and PROBA-3. His research aims at enabling future miniature distributed space systems for unprecedented remote sensing, space and planetary science, exploration and spaceflight sustainability. He performs fundamental and applied research at the intersection of advanced astrodynamics, spacecraft Guidance, Navigation and Control (GNC), autonomy, decision making and space system engineering. Dr. D'Amico is institutional PI of three autonomous satellite swarm missions funded by NASA and NSF, namely STARLING, VISORS, and SWARM-EX. He is Fellow of AAS, Associate Fellow of AIAA, Associate Editor of AIAA JGCD, Advisor of NASA and three space startups (Capella, Infinite Orbits, Reflect Orbital). He was the recipient of several awards, including Best Paper Awards at IAF (2022), IEEE (2021), AIAA (2021), AAS (2019) conferences, the Leonardo 500 Award by the Leonardo da Vinci Society/ISSNAF (2019), FAI/NAA's Group Diploma of Honor (2018), DLR's Sabbatical/Forschungssemester (2012) and Wissenschaft Preis (2006), and NASA's Group Achievement Award for the GRACE mission (2004).
\end{biographywithpic}

\begin{biographywithpic}
{Marco Pavone}{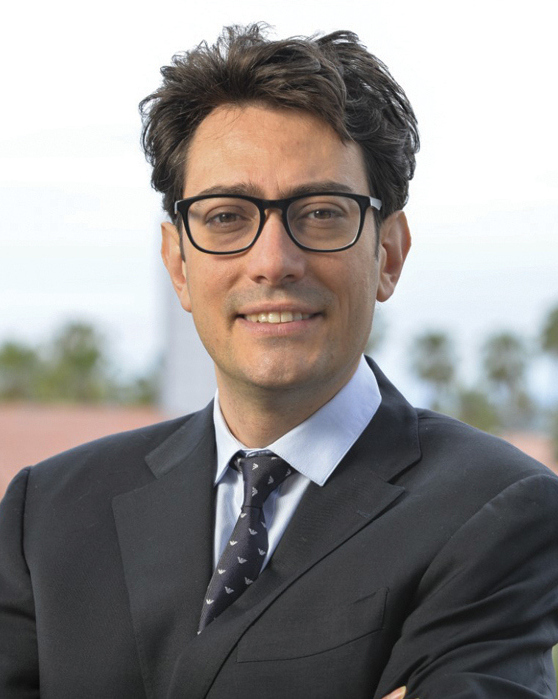}
Dr. Marco Pavone is an Associate Professor of Aeronautics and Astronautics at Stanford University, where he is the Director of the Autonomous Systems Laboratory and Co-Director of the Center for Automotive Research at Stanford. Before joining Stanford, he was a Research Technologist within the Robotics Section at the NASA Jet Propulsion Laboratory. He received a Ph.D. degree in Aeronautics and Astronautics from the Massachusetts Institute of Technology in 2010. His main research interests are in the development of methodologies for the analysis, design, and control of autonomous systems, with an emphasis on self-driving cars, autonomous aerospace vehicles, and future mobility systems. He is a recipient of a number of awards, including a Presidential Early Career Award for Scientists and Engineers from President Barack Obama, an Office of Naval Research Young Investigator Award, a National Science Foundation Early Career (CAREER) Award, a NASA Early Career Faculty Award, and an Early-Career Spotlight Award from the Robotics Science and Systems Foundation. He was identified by the American Society for Engineering Education (ASEE) as one of America's 20 most highly promising investigators under the age of 40. His work has been recognized with best paper nominations or awards at the European Control Conference, at the IEEE International Conference on Intelligent Transportation Systems, at the Field and Service Robotics Conference, at the Robotics: Science and Systems Conference, at the ROBOCOMM Conference, and at NASA symposia. He is currently serving as an Associate Editor for the IEEE Control Systems Magazine. He is serving or has served on the advisory board of a number of autonomous driving start-ups (both small and multi-billion dollar ones), he routinely consults for major companies and financial institutions on the topic of autonomous systems, and is a venture partner for investments in AI-enabled robots.
\end{biographywithpic}

\end{document}